\documentclass[sigconf]{acmart}

\AtBeginDocument{%
  \providecommand\BibTeX{{%
    \normalfont B\kern-0.5em{\scshape i\kern-0.25em b}\kern-0.8em\TeX}}}

\copyrightyear{2021}
\acmYear{2021}
\setcopyright{acmcopyright}\acmConference[KDD '21]{Proceedings of the 27th ACM
SIGKDD Conference on Knowledge Discovery and Data Mining}{August 14--18,
2021}{Singapore, Singapore}
\acmBooktitle{Proceedings of the 27th ACM SIGKDD Conference on Knowledge
Discovery and Data Mining (KDD '21), August 14--18, 2021, Singapore, Singapore}
\acmPrice{15.00}
\acmDOI{10.1145/1122445.1122456}
\acmISBN{978-1-4503-XXXX-X/21/08}

\setcopyright{acmcopyright}
\copyrightyear{2021}
\acmYear{2021}
\acmDOI{10.1145/1122445.1122456}

\usepackage[ruled, lined, vlined, linesnumbered]{algorithm2e}
\usepackage{amsmath}
\usepackage{amsthm}
\usepackage{booktabs}       
\usepackage{multirow}
\usepackage{nccmath}
\usepackage[caption=false,font=normalsize]{subfig}
\usepackage{url}            
\usepackage{xcolor}
\usepackage{xspace}

\theoremstyle{definition}
\newtheorem{definition}{Definition}

\newtheorem{theorem}{Theorem}
\newtheorem{lemma}{Lemma}

\DeclareMathOperator*{\argmin}{arg\,min}

\newcommand{\eg}{\emph{e.g.},\xspace}
\newcommand{\ie}{\emph{i.e.},\xspace}

\newcommand\figref[1]{Fig.~\ref{#1}}
\newcommand\tabref[1]{Table~\ref{#1}}
\newcommand\secref[1]{Sec.~\ref{#1}}

\newcommand\appref[1]{Appendix~\ref{#1}}

\newcommand{\fakeparagraph}[1]{\vspace{1mm}\noindent\textbf{#1.}}

\ifodd 1

\else

\fi

\newcommand{\sysname}{PAM\xspace}

\begin{document}
\fancyhead{}

\title{Pruning-Aware Merging for Efficient Multitask Inference}

\author{Xiaoxi He}
\affiliation{%
  \institution{ETH Zürich}
  \city{Zürich}
  \country{Switzerland}}
\email{hex@ethz.ch}

\author{Dawei Gao}
\affiliation{%
  \institution{SKLSDE \& BDBC, Beihang University}
  \city{Beijing}
  \country{China}}
\email{david_gao@buaa.edu.cn}

\author{Zimu Zhou}
\affiliation{%
  \institution{Singapore Management University}
  \city{Singapore}
  \country{Singapore}}
\email{zimuzhou@smu.edu.sg}

\author{Yongxin Tong}
\affiliation{%
  \institution{SKLSDE \& BDBC, Beihang University}
  \city{Beijing}
  \country{China}}
\email{yxtong@buaa.edu.cn}

\author{Lothar Thiele}
\affiliation{%
  \institution{ETH Zürich}
  \city{Zürich}
  \country{Switzerland}}
\email{thiele@ethz.ch}



\begin{abstract}
Many mobile applications demand selective execution of multiple correlated deep learning inference tasks on resource-constrained platforms.
Given a set of deep neural networks, each pre-trained for a single task, it is desired that executing arbitrary combinations of tasks yields minimal computation cost.
Pruning each network separately yields suboptimal computation cost due to task relatedness.
A promising remedy is to merge the networks into a multitask network to eliminate redundancy across tasks before network pruning.
However, pruning a multitask network combined by existing network merging schemes cannot minimise the computation cost of every task combination because they do not consider such a future pruning. 
To this end, we theoretically identify the conditions such that pruning a multitask network minimises the computation of all task combinations.
On this basis, we propose Pruning-Aware Merging (\sysname), a heuristic network merging scheme to construct a multitask network that approximates these conditions.
The merged network is then ready to be further pruned by existing network pruning methods.
Evaluations with different pruning schemes, datasets, and network architectures show that \sysname achieves up to $4.87\times$ less computation against the baseline without network merging, and up to $2.01\times$ less computation against the baseline with a state-of-the-art network merging scheme.
\end{abstract}

\maketitle

\section{Introduction}
\label{sec:introduction}

Deep neural networks that can run locally on resource-constrained devices hold potential for various emerging applications such as autonomous drones and social robots \cite{bib:MobiCom18:Fang, bib:MobiSys20:Lee}. 
These applications often simultaneously perform a set of correlated inference tasks based on the current context to deliver accurate and adaptive services.
Although deep neural networks pre-trained for individual tasks are readily available \cite{bib:PIEEE98:LeCun, bib:arXiv14:Simonyan}, deploying multiple such networks easily overwhelms the resource budget.

To support these applications on low-resource platforms, we investigate \textit{efficient multitask inference}.
Given a set of correlated inference tasks and deep neural networks (each network pre-trained for an individual task), we aim to minimise the computation cost when \textbf{any subset of tasks} is performed at inference time.

One naive solution to efficient multitask inference is to \textit{prune} each network for individual tasks \textit{separately}.
A deep neural network is typically over-parameterised \cite{bib:NIPS13:Denil}.
Network pruning \cite{bib:ICML18:Dai, bib:PIEEE20:Deng, bib:KDD20:Gao, bib:CVPR19:Molchanov, bib:PIEEE17:Sze} can radically reduce the number of operations within a network without accuracy loss in the inference task.
This solution, however, is only optimal if \textit{a single task} is executed at a time.
When multiple correlated tasks are running concurrently, this solution is unable to save computation cost by exploiting tasks relatedness and sharing intermediate results among networks.

\begin{figure}[t]
\centering
\includegraphics[width=0.9\linewidth]{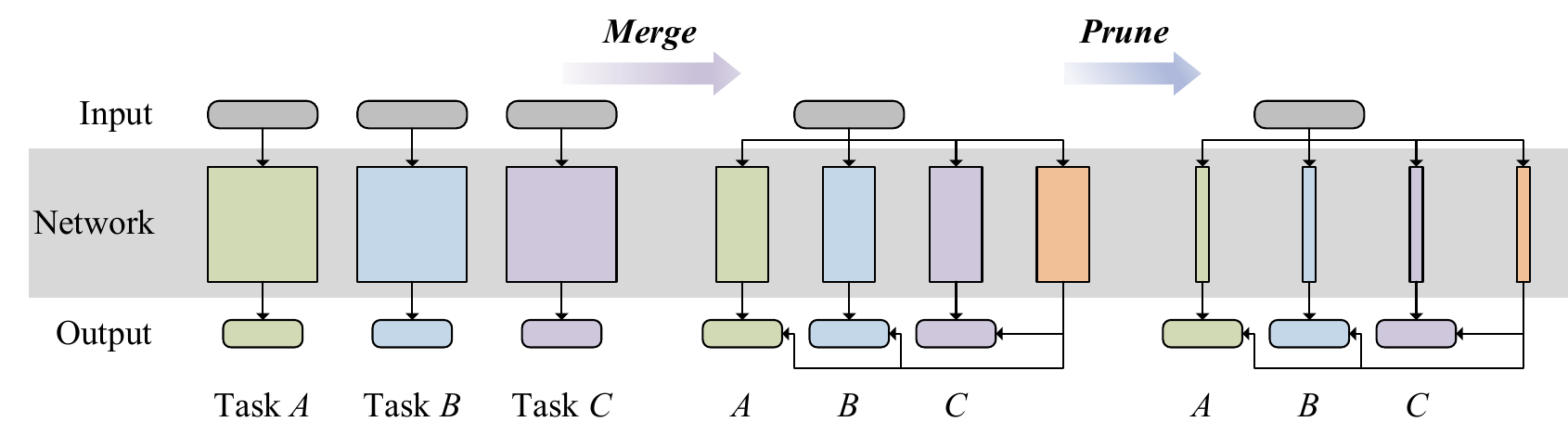}
\caption{Efficient multitask inference by ``merge \& prune''. Three networks pre-trained for tasks $A$, $B$ and $C$ are first merged into a multitask network and then pruned.}
\label{fig:intro}
\end{figure}

A more promising solution framework is \textit{``merge \& prune''}, which merges multiple networks into a \textit{multitask network}, before pruning it (\figref{fig:intro}).
A few pioneer studies \cite{bib:IJCAI18:Chou, bib:NIPS18:He} have explored network merging schemes to eliminate the redundancy among multiple networks pre-trained for correlated tasks.
However, pruning a multitask network merged via these schemes can only minimise computation cost when \textit{all tasks} are executed at the same time.

In this paper, we propose Pruning-Aware Merging (\sysname), a new network merging scheme for efficient multitask inference.
By applying existing network pruning methods on the multitask network merged by \sysname, the computation cost when performing \textit{any subset} of tasks can be reduced.
Extensive experiments show that ``PAM \& Prune'' consistently achieves solid advantages over the state-of-the-art network merging scheme across tasks, datasets, network architectures and pruning methods.

Our main contributions and results are as follows:
\begin{itemize}
    \item
    We theoretically show that pruning a multitask network may not simultaneously minimise the computation cost of all task combinations in the network.
    We then identify conditions such that minimising the computation of all task combinations via network pruning becomes feasible.
    To the best of our knowledge, this is the first explicit analysis on the applicability of network pruning in multitask networks.
    \item
    We propose Pruning-Aware Merging (\sysname), a heuristic network merging scheme to construct a multitask network that approximately meets the conditions in our analysis and enables ``merge \& prune'' for efficient multitask inference.
    \item 
    We evaluate \sysname with various pruning schemes, datasets and architectures.
    \sysname achieves up to $4.87\times$ less computation cost against the baseline without network merging, and up to $2.01\times$ less computation cost against the baseline with the state-of-the-art network merging scheme \cite{bib:NIPS18:He}.
\end{itemize}

In the rest of this paper, we review related work in \secref{sec:related}, introduce our problem statement in \secref{sec:notation}, theoretical analysis in \secref{sec:redundancy} and our solution in \secref{sec:method}.
We present the evaluations of our methods in \secref{sec:experiments} and finally conclude in \secref{sec:conclusion}.
\section{Related Work}
\label{sec:related}
Our work is related to the following categories of research.

\fakeparagraph{Network Pruning}
Network pruning reduces the number of operations in a deep neural network without loss in accuracy \cite{bib:PIEEE20:Deng, bib:PIEEE17:Sze}.
Unstructured pruning removes unimportant weights \cite{bib:NIPS17:Dong, bib:KDD20:Gao, bib:NIPS16:Guo}.
However, customised hardware \cite{bib:ISCA16:Han} is compulsory to exploit such irregular sparse connections for acceleration.
Structured pruning enforces sparsity at the granularity of channels/filters/neurons \cite{bib:ICML18:Dai, bib:ICLR17:Li, bib:CVPR19:Molchanov, bib:NIPS16:Wen}.
The resulting sparsity is fit for acceleration on general-purpose processors.
Prior pruning proposals implicitly assume a single task in the given network.
We identify the challenges to prune a multitask network and propose a network merging scheme such that pruning the merged multitask network minimises computation cost of all task combinations in the network.

\fakeparagraph{Multitask Networks}
A multitask network can be either constructed from scratch via Multi-Task Learning (MTL) or merged from multiple networks pre-trained for individual tasks.
MTL joint trains multiple tasks for better generalisation \cite{bib:NSR17:Zhang}, while we focus on the computation cost of running multiple tasks at inference time.
Network merging schemes \cite{bib:IJCAI18:Chou, bib:NIPS18:He} aim to construct a compact multitask network from networks pre-trained for individual tasks.
Both MTZ \cite{bib:NIPS18:He} and NeuralMerger \cite{bib:IJCAI18:Chou} enforce weight sharing among networks to reduce their overall storage.
In contrast, we account for the computation cost of a multitask network.
Although constructing a multitask network using these schemes \cite{bib:IJCAI18:Chou, bib:NIPS18:He} and pruning it via existing pruning methods can reduce the computation when \textit{all tasks} are concurrently executed, they cannot minimise the computation cost for \textit{every combination of tasks}.
\section{Problem Statement}
\label{sec:notation}

\begin{figure}[t]
\centering
\includegraphics[width=0.95\linewidth]{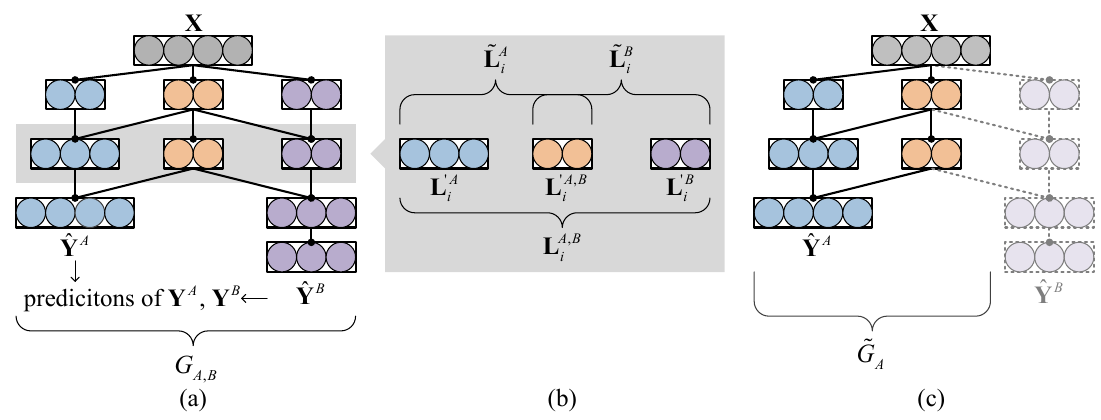}
\caption{Important notations: (a) graph representation $G_{A,B}$ of a multitask network for tasks $A$ and $B$, with $N_{A}=2$ hidden layers for task $A$ and $N_{B}=3$ hidden layers for task $B$; (b) layer outputs for the $i$-th layer; (c) subgraph $\widetilde{G}_A$ for task $A$.}
\label{fig:illustration}
\end{figure}

We define and analyse our problem based on the graph representation of neural networks.
The graph representation reflects the computation cost of neural networks (see below) and facilitates an information theoretical understanding on network pruning (see \secref{sec:redundancy}).
\figref{fig:illustration} shows important notations used throughout this paper.
For ease of illustration, we explain our analysis using two tasks.
Extensions to more than two tasks are in \secref{appendix:extension}.

\subsection{Graph Representation of Neural Networks}
\fakeparagraph{Task}
Consider three sets of random variable $\mathbf{X}\in\mathcal{X}$, $\mathbf{Y}^A\in\mathcal{Y}^A$, and $\mathbf{Y}^B\in\mathcal{Y}^B$.
Task $A$ outputs $\widehat{\mathbf{Y}}^A$, a prediction of $\mathbf{Y}^A$, by learning the conditional distribution $\text{Pr}\{\mathbf{Y}^A=\mathbf{y}|\mathbf{X}=\mathbf{x}\}$. 
Task $B$ outputs $\widehat{\mathbf{Y}}^B$, a prediction of $\mathbf{Y}^B$, by learning $\text{Pr}\{\mathbf{Y}^B=\mathbf{y}|\mathbf{X}=\mathbf{x}\}$.

\fakeparagraph{Single-Task Network}
For task $A$, a neural network without feedback loops can be represented by an acyclic directed graph $G_A=\{V^A,E^A\}$.
Each vertex represents a neuron.
There is an edge between two vertices if two neurons are connected.
The vertex set $V_A$ can be categorised into three types of nodes: source, internal and sink node.
$\text{deg}^-(v)$/$\text{deg}^+(v)$ is the indegree/outdegree of a vertex $v$.
\begin{itemize}
\item
Source node set $\mathbf{v}^A_X = \{v|v\in V^A \wedge \text{deg}^-(v)=0\}$ represents the \textit{input layer}.
Each source node represents an \textit{input neuron} and outputs a random variable $X_i\in\mathbf{X}$. 
The output of the input layer is the input random variable set $\mathbf{X}$.
\item
Internal nodes $v_i \in \{v|v\in V \wedge\ \text{deg}^-(v)\neq 0 \wedge \text{deg}^+(v)\neq 0\}$ represents the \textit{hidden neurons}. 
The output of each hidden neuron is generated by calculating the weighted sum of its inputs and then applying an activation function.
\item
Sink node set $\mathbf{v}^A_Y = \{v|v\in V \wedge \text{deg}^+(v)=0\}$ represents the \textit{output layer}. 
Each sink node represents an \textit{output neuron} and the output is calculated in the same way as the hidden neurons. 
The output of the output layer is the prediction $\widehat{\mathbf{Y}}^A$ of ground-truth labels $\mathbf{Y}^A$.
\end{itemize}

We organise the hidden neurons $v_i$ of $G^A$ into layers $\mathbf{v}_i^A$ by Algorithm \ref{alg:layer_construct}.
$N^+(\mathbf{v})$ represents the out-coming neighbours of the vertex set $\mathbf{v}$.
Algorithm \ref{alg:layer_construct} can organise any acyclic single-task network into layers and the layer outputs satisfy the Markov property.

\begin{algorithm}[t]
    \SetAlgoLined
    \KwIn{A neural network graph $G^A$}
    \KwOut{$N+1$ layers $\mathbf{v}_i^A$ with $i=1,\cdots, N+1$.}
	\BlankLine
	
	$\mathbf{v}_0^A \gets \mathbf{v}_X^A $\; 
	
	$i \gets 0$\;
	
	\While{$N^+(\mathbf{v}_i^A)\neq \mathbf{v}_Y^A$}{
	    $\mathbf{v}_{i+1}^A \gets \emptyset$\;
	    
	    \For{each node $v_{i,j}^A \in \mathbf{v}_i^A$}{
	        \If{$N^+(v_{i,j}^A) \cap \mathbf{v}_Y^A \neq \emptyset$}{
	            $\mathbf{v}_{i+1}^A \gets \mathbf{v}_{i+1}^A \cup \{v_{i,j}^A\}$\;
	            
	        }
	    }
	    
	    $\mathbf{v}_{i+1}^A \gets \mathbf{v}_{i+1}^A \cup \big( N^+(\mathbf{v}_i^A)\setminus \mathbf{v}_i^A \big)$\;
	    
	    $i \gets i+1$\;
	}
	$N \gets i$\;
	
	$\mathbf{v}_{N+1}^A \gets \mathbf{v}_Y^A$\;
	
	\caption{Organise vertices in the graph representation of a neural network into layers.}
	\label{alg:layer_construct}
\end{algorithm}

\fakeparagraph{Multitask Network}
For task $A$ and $B$, a multitask network without feedback loops can be represented by an acyclic directed graph $G_{A,B}$.
All paths from the input neurons to the output neurons for task $A$ form a subgraph $\widetilde{G}_A$ (see \figref{fig:illustration}(c)), which is in effect the same as a single-task network. 
When only task $A$ is performed, only $\widetilde{G}_A$ is activated. 
Subgraph $\widetilde{G}_B$ is defined similarly. 
We also organise vertices of $\widetilde{G}_A$ and $\widetilde{G}_B$ into layers with Algorithm \ref{alg:layer_construct}.
Layer outputs of $\widetilde{G}_A$ and $\widetilde{G}_B$ are denoted as $\widetilde{\mathbf{L}}_i^A$ and $\widetilde{\mathbf{L}}_i^B$.
Suppose $\widetilde{G}_A$ and $\widetilde{G}_B$ have respectively $N_A$ and $N_B$ hidden layers.
We assume $N_A \leq N_B$ w.l.o.g..
Then the $i$-th layer output of $G_{A,B}$ is defined as $\mathbf{L}_i^{A,B} = \widetilde{\mathbf{L}}_i^A \cup \widetilde{\mathbf{L}}_i^B$ with $i = 0,\cdots,N_A$. 
As shown in \figref{fig:illustration}(b), $\mathbf{L}_i^{A,B}$ consists of three sets of neurons: $\mathbf{L}'^A_i$, $\mathbf{L}'^B_i$ and $\mathbf{L}'^{A,B}_i$.

\fakeparagraph{Remarks}
The above definitions have two benefits.
\textit{(i)}
The computation cost of a neural network is an \textit{increasing function} of the size of the graph, \ie the number of edges plus vertices.
Reducing the computation cost of the network is transformed into removing edges or vertices in the graph.
\textit{(ii)}
For a single-task network with $N_A$ hidden layers, its layer outputs form a Markov chain: $\mathbf{Y}^A \to \mathbf{L}_0^A \to \cdots \to \mathbf{L}_{N_A+1}^A$.
All layer outputs $\mathbf{L}_i^{A,B}$ in a multitask network also form a Markov chain.
The Markov property allows an information theoretical analysis on neural networks \cite{bib:ICLR18:Saxe,bib:ITW15:Tishby}.

\subsection{Problem Definition} 
\label{subsec:problem_definition}
Given two single-task networks $G_A$ and $G_B$ pre-trained for task $A$ and $B$, we aim to construct a multitask network $G_{A,B}$ such that pruning on $G_{A,B}$ can minimise the number of vertices and edges in $G_{A,B}$, $\widetilde{G}_A$ and $\widetilde{G}_B$ while preserving inference accuracy on $A$ and $B$.
To ensure minimal computation of \textit{any subset} of tasks, we need to minimise the number of vertices and edges in \textit{any subgraph}.
For two tasks, $G_{A,B}$ corresponds to running task $A$ and $B$ concurrently; $\widetilde{G}_A$ ($\widetilde{G}_B$) corresponds to running task $A$ ($B$) only.
Next, we show the difficulty to optimise all subgraphs simultaneously.

\section{Theoretical Understanding}
\label{sec:redundancy}
This section presents a theoretical understanding on the challenges to prune a multitask network and identifies conditions such that minimising the computation cost of all task combinations via pruning becomes feasible (Theorem \ref{theorem:PAM}).
Proofs are in \appref{appendix:proofs}.

\subsection{Why Pruning a Single-task Network Work}
\label{subsec:redundancy_single}

Pruning a single-task network reduces the computation cost of a neural network while retaining task inference accuracy by suppressing \textit{redundancy} in the network \cite{bib:PIEEE20:Deng, bib:PIEEE17:Sze}.
From the information theoretical perspective \cite{bib:ICLR18:Saxe,bib:ITW15:Tishby}, 
since the layer outputs form a Markov chain, the inference accuracy for a given task $A$ is positively correlated to the task related information transmitted through the network at each layer, measured by $I(\mathbf{L}^A_i;\mathbf{Y}^A)$.
All other information is irrelevant for the task.
Hence \textit{the redundancy within a single-task network} can be defined as below.
\begin{definition}\label{def:single}
For the $i$-th layer in the single-task neural network $G_A$, the redundancy of the layer is defined as $\mathcal{R}_A(\mathbf{L}^A_i) = \sum_{L^A_{i,j}\in \mathbf{L}^A_i} H(L^A_{i,j})-I(\mathbf{L}^A_i;\mathbf{Y}^A)$.
\end{definition}
$\sum_{L^A_{i,j}\in \mathbf{L}^A_i}H(L^A_{i,j})$ measures \textit{the maximal amount of information} the layer can express.
$I(\mathbf{L}^A_i;\mathbf{Y}^A)$ measures \textit{the amount of task $A$ related information} in the layer output.
By definition, $\mathcal{R}_A(\mathbf{L}^A_i) \geq 0$. 

\fakeparagraph{Remarks}
$\sum_{L^A_{i,j}\in \mathbf{L}^A_i}H(L^A_{i,j})$ is positively correlated to the number of vertices and incoming edges of the $i$-th layer. 
Therefore, in a well trained network where $I(\mathbf{L}^A_i;\mathbf{Y}^A)$ can no longer increase, the computation cost can be minimised by reducing $\mathcal{R}_A(\mathbf{L}^A_i)$.

Accordingly, pruning a single-task network can be formalised as an optimisation problem
\begin{equation}
    \label{eqt:single_pruning}
    \text{minimise } \medop{\sum_{i=1}^{N_A+1}} \left( \mathcal{R}_A(\mathbf{L}_i^A) - \xi_i \cdot I(\mathbf{L}_i^A;\mathbf{Y}^A)\right)
\end{equation}
where $\xi_i > 0$ controls the trade-off between inference accuracy and computation cost.

\fakeparagraph{Remarks}
Existing pruning methods implicitly assume a single-task network.
That is, they are all designed to solve optimisation problem \eqref{eqt:single_pruning}, even though the concrete strategies vary.
We now show the problems that occur when these pruning methods are applied to a multitask network.

\subsection{Why Pruning a Multitask Network Fail}
\label{subsec:redundancy_multi}
As mentioned in \secref{subsec:problem_definition}, we aim to minimise the computation cost of any subset of tasks, which is a \textit{multi-objective} optimisation problem.
As we will show below, existing network pruning methods are unable to handle these objectives simultaneously.

We first define redundancy when performing two tasks at the same time, similarly as in Definition \ref{def:single}.
\begin{definition}
\label{def:multi}
For a multitask network $G_{A,B}$, the redundancy of its $i$-th layer is $\mathcal{R}_{A,B}(\mathbf{L}^{A,B}_i) = \sum_{L^{A,B}_{i,j}\in \mathbf{L}^{A,B}_i}H(L^{A,B}_{i,j})-I(\mathbf{L}^{A,B}_i;\mathbf{Y}^A, \mathbf{Y}^B)$.
\end{definition}

Following the above definitions of redundancy, our objective in \secref{subsec:problem_definition} is equivalent to minimising the redundancy in $G_{A,B}$ as well as in its two subgraphs $\widetilde{G}_A$ and $\widetilde{G}_B$, which leads to the following three-objective optimisation (still, we assume $N_A \leq N_B$ w.l.o.g.):
\begin{equation}
    \label{eqt:multi_pruning}
    \begin{aligned}
    & \text{minimise } \\
    & \medop{\sum_{i=1}^{N_A+1}} \left( \mathcal{R}_A(\widetilde{\mathbf{L}}_i^A) - \tilde{\xi}^A_i \cdot I(\widetilde{\mathbf{L}}_i^A;\mathbf{Y}^A)\right), \\
    &  \medop{\sum_{i=1}^{N_B+1}} \left( \mathcal{R}_B(\widetilde{\mathbf{L}}_i^B) - \tilde{\xi}^B_i \cdot I(\widetilde{\mathbf{L}}_i^B;\mathbf{Y}^B)\right),\\ 
     &  \medop{\sum_{i=1}^{N_A}} \left( \mathcal{R}_{A,B}(\mathbf{L}_i^{A,B}) - \xi^A_i \cdot I(\widetilde{\mathbf{L}}_i^A;\mathbf{Y}^A) - \xi^B_i \cdot I(\widetilde{\mathbf{L}}_i^B;\mathbf{Y}^B)\right)\\
    \end{aligned}
\end{equation}
Reducing $\mathcal{R}_A(\widetilde{\mathbf{L}}_i^A)$, $\mathcal{R}_B(\widetilde{\mathbf{L}}_i^B)$ and $\mathcal{R}_{A,B}(\mathbf{L}_i^{A,B})$ decreases the number of vertices and edges in $\widetilde{G}_A$, $\widetilde{G}_B$ and $G_{A,B}$, respectively.
$\xi^A_i, \xi^B_i, \tilde{\xi}^A_i, \tilde{\xi}^B_i > 0$ are parameters to control the trade-off between computation cost and inference accuracy, as well as to balance task $A$ and $B$.

To solve optimisation problem \eqref{eqt:multi_pruning} with prior network pruning methods, we observe two problems.

\fakeparagraph{Problem 1: The first two objectives in \eqref{eqt:multi_pruning} may conflict}
This is because reducing $\mathcal{R}_B(\widetilde{\mathbf{L}}_i^B)$ may decrease $I(\widetilde{\mathbf{L}}_i^A;\mathbf{Y}^A)$ (proofs in \appref{appendix:problem_1}).
In other words, when pruning subgraph $\widetilde{G}_B$, it is possible that some information related to task A is removed from the shared vertices between $\widetilde{G}_A$ and $\widetilde{G}_B$. 
Hence $I(\widetilde{\mathbf{L}}_i^A;\mathbf{Y}^A)$ decreases and the inference accuracy of task $A$ deteriorates. 

\fakeparagraph{Problem 2: It is unclear how to minimise the third objective in \eqref{eqt:multi_pruning}}
As mentioned in \secref{subsec:redundancy_single}, most pruning methods are designed with a single-task network in mind.
It is unknown how to apply them to a multitask network $G_{A,B}$ with architecture in \figref{fig:illustration} (a).

\subsection{When Pruning a Multitask Network Work}
\label{subsec:conditions}
The two problems in \secref{subsec:redundancy_multi} show that not all multitask networks can be pruned for efficient multitask inference.
However, a multitask network can be effectively pruned if it meets the conditions stated by the following theorem.
\begin{theorem}
    \label{theorem:PAM}
     If $\forall\, 1 \le i \le N_A$, the conditions below are satisfied:
     \begin{equation}
     \label{eqt:conditions}
     \begin{aligned}
        I(\mathbf{L}_i'^A;\mathbf{L}_i'^B;\mathbf{Y}^A;\mathbf{Y}^B) = 0 \\ I(\mathbf{L}_i'^{A,B};\mathbf{Y}^A|\mathbf{L}_i'^A,\mathbf{Y}^B) = 0 \\ I(\mathbf{L}_i'^{A,B};\mathbf{Y}^B|\mathbf{L}_i'^B,\mathbf{Y}^A) = 0
    \end{aligned}
     \end{equation}
    where $I(\mathbf{L}_i'^A;\mathbf{L}_i'^B;\mathbf{Y}^A;\mathbf{Y}^B)$ is the \textit{co-information}~\cite{bib:ICA03:Bell}, then the \textit{three-objective} optimisation problem \eqref{eqt:multi_pruning} can be reduced to \textit{two} non-conflicting optimisation problems that can be solved \textit{independently}:
    \begin{equation}
    \label{eqt:2opt}
    \begin{aligned}
    \text{minimise } \medop{\sum_{i=1}^{N_A+1}}  \mathcal{R}_A(\widetilde{\mathbf{L}}_i^A) - \tilde{\xi}^A_i \cdot I(\widetilde{\mathbf{L}}_i^A;\mathbf{Y}^A), \\
     \text{minimise }  \medop{\sum_{i=1}^{N_B+1}}  \mathcal{R}_B(\widetilde{\mathbf{L}}_i^B) - \tilde{\xi}^B_i \cdot I(\widetilde{\mathbf{L}}_i^B;\mathbf{Y}^B)
     \end{aligned}
    \end{equation}
\end{theorem}
Each of the two optimisation problems \eqref{eqt:2opt} are in effect single-task pruning problem like optimisation problem \eqref{eqt:single_pruning}, which can be effectively solved by prior pruning proposals.

\fakeparagraph{Remarks}
Theorem~\ref{theorem:PAM} provides important guidelines to design the network merging scheme for our problem in \secref{subsec:problem_definition}.
Specifically, if $G_A$ and $G_B$ can be merged into a a multitask network $G_{A,B}$ such that conditions \eqref{eqt:conditions} are satisfied, we can simply apply existing network pruning on the two subgraphs $\widetilde{G}_A$ and $\widetilde{G}_B$ to minimise the computation cost when performing any subset of tasks.

\section{Pruning-Aware Merging}
\label{sec:method}
Based on the above analysis, we propose Pruning-Aware Merging (\sysname), a novel network merging scheme that constructs a multitask network from pre-trained single task networks.
\sysname approximately meets the conditions in Theorem~\ref{theorem:PAM} such that the merged multitask network can be effectively pruned for efficient multitask inference.

\begin{figure*}[t]
\centering
\includegraphics[width=0.70\linewidth]{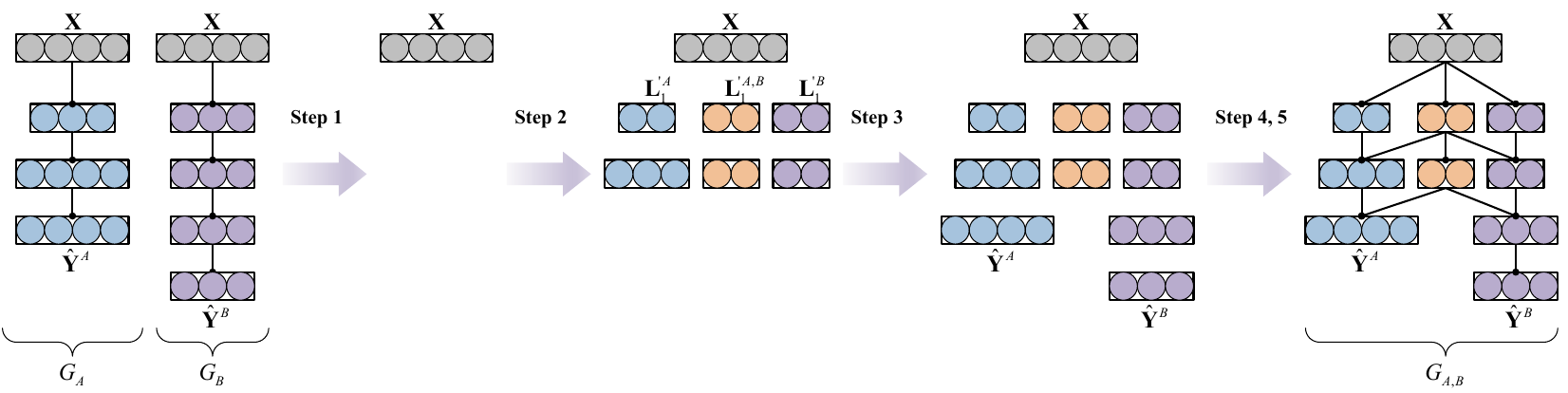}
\caption{\sysname workflow to construct a multitask network ($G_{A,B}$) from two single-task networks ($G_{A}$ and $G_{B}$).}
\label{fig:workflow}
\end{figure*}

\subsection{\sysname Workflow}
\label{subsec:framework}
Given two single-task networks $G_A$ and $G_B$ pre-trained for task $A$ and $B$ ($N_A \leq N_B$), \sysname constructs a multitask network $G_{A,B}$ with the steps below (see \figref{fig:workflow}).
\begin{enumerate}
    \item 
    Assign $\mathbf{L}^{A,B}_0 = \mathbf{X}$, as $G_{A,B}$, $G_A$ and $G_B$ use the same inputs.
    \item 
    For $i = 1,\cdots,N_A$, regroup the neurons from $\mathbf{L}_i^A$ and $\mathbf{L}_i^B$ into $\mathbf{L}'^A_i$, $\mathbf{L}'^B_i$ and $\mathbf{L}'^{A,B}_i$ by the regrouping algorithm in \secref{subsec:algorithm}.
    \item 
    Take over the output layer for task $A$: $\widetilde{\mathbf{L}}^A_{N_A+1} = \mathbf{L}^A_{N_A+1}$.
    For $i = N_A+1,\cdots,N_B+1$, take over the remaining layers from $G_B$: $\widetilde{\mathbf{L}}^B_i = \mathbf{L}^B_i$.
    \item 
    Reconnect the neurons as in \figref{fig:workflow}.
    If a connection exist before merging, it preserves its original weight. 
    Otherwise it is initialised with a zero.
    \item 
    Finetune $G_{A,B}$ on $A$ and $B$ to learn the newly added connections. 
    For the shared connections, $\mathbf{L}_{i-1}'^{A,B} \to \mathbf{L}_i'^{A,B}$. 
    The gradients are first calculated separately on $A$ and $B$, and then averaged before weight updating.
\end{enumerate}
Now the multitask network $G_{A,B}$ is ready to be pruned. 
From Theorem \ref{theorem:PAM}, we can apply network pruning on the two subgraphs $\widetilde{G}_A$ and $\widetilde{G}_B$ independently and achieve a minimal computation cost for all combinations of tasks.
However, since we only approximate the conditions in \eqref{eqt:conditions}, pruning $\widetilde{G}_A$ and $\widetilde{G}_B$ is not perfectly independent in practice. 
Hence we prune $\widetilde{G}_A$ and $\widetilde{G}_B$ \textit{in an alternating manner} to balance between task $A$ and $B$.

\subsection{Regrouping Algorithm}
\label{subsec:algorithm}

The core of \sysname is the regrouping algorithm in the second step in \secref{subsec:framework}. 
It regroups the neurons from $\mathbf{L}_i^A$ and $\mathbf{L}_i^B$ into three sets: $\mathbf{L}'^A_i$, $\mathbf{L}'^B_i$ and $\mathbf{L}'^{A,B}_i$, such that the conditions \eqref{eqt:conditions} in Theorem \ref{theorem:PAM} are satisfied.
However, it is computation-intensive to estimate the co-information and conditional mutual information in \eqref{eqt:conditions} precisely.
We rely on the following theorem to approximate the conditions.
\begin{theorem} \label{theorem:approx}
    The conditions in \eqref{eqt:conditions} can be achieved by minimising $I(\mathbf{L}_i'^A;\mathbf{Y}^B)$, $I(\mathbf{L}_i'^B;\mathbf{Y}^A)$, and maximising $I(\mathbf{L}_i'^A;\mathbf{Y}^A)$, $I(\mathbf{L}_i'^B;\mathbf{Y}^B)$.
\end{theorem}
\fakeparagraph{Remarks}
$I(\mathbf{L}_i'^A;\mathbf{Y}^B)$ and $I(\mathbf{L}_i'^B;\mathbf{Y}^A)$ describe the ``misplaced'' information, i.e., the information that is useful for one task, but contained in neurons that are not connected to the outputs of this task.
Therefore such information is redundant and needs to be minimised.
$I(\mathbf{L}_i'^A;\mathbf{Y}^A)$ and $I(\mathbf{L}_i'^B;\mathbf{Y}^B)$ measure the ``relevant'' information, i.e., the information useful for one task and contained in neurons connected to this task.
Note that this information may not be simply maximised, because it includes the information that is useful for both tasks.
It requires simultaneously minimising the ``misplaced'' information and maximising the ``correct'' information to achieve the conditions in \eqref{eqt:conditions}.
The proof of Theorem~\ref{theorem:approx} is in \secref{appendix:proof_2}.

\begin{algorithm}[t]
	\SetAlgoLined
	\KwIn{$\mathbf{L}_i^A$, $\mathbf{L}_i^B$, $\mathbf{X}$, $\mathbf{Y}^A$, $\mathbf{Y}^B$, $\alpha$}
	\KwOut{$\mathbf{L}_i'^A$, $\mathbf{L}_i'^B$, $\mathbf{L}_i'^{A,B}$}
	\BlankLine
	$N = \min\{N^A, N^B\}$\;
	
	\For{$i\gets 1$ \KwTo $N$ }{
		$\mathbf{F}^A \gets \mathbf{F}^B \leftarrow \mathbf{L}_i^A \cup \mathbf{L}_i^B $\; 
		
		$\mathbf{L}_i'^A \leftarrow \emptyset $\;

		\While{$I(\mathbf{L}_i'^A;\mathbf{Y}^B)\leq \alpha$}{
			$L_{i,\cdot} \gets \argmin_{L_{i,j} \in \mathbf{F}_i^A} I( \{L_{i,j}\}\cup \mathbf{L}_i'^A; \mathbf{Y}^B)$\; 
			
			move the neuron $L_{i,\cdot}$ from $\mathbf{F}^A$ to $\mathbf{L}_i'^A$
			
		}
		$\mathbf{L}_i'^B\leftarrow \emptyset $ \;

		\While{$I(\mathbf{L}_i'^B;\mathbf{Y}^A)\leq \alpha$}{
			$L_{i,\cdot} \gets \argmin_{L_{i,j}\in \mathbf{F}_i^B} I(\{L_{i,j}\} \cup \mathbf{L}_i'^B; \mathbf{Y}^A)$\;
	 
			move the neuron $L_{i,\cdot}$ from $\mathbf{F}^B$ to $\mathbf{L}_i'^B$
		}
	The remaining neurons join $\mathbf{L}_i'^{A,B}$: $\mathbf{L}_i'^{A,B} \gets \mathbf{L}_i^A \cup \mathbf{L}_i^B \setminus \Big(\mathbf{L}_i'^A \cup \mathbf{L}_i'^B \Big)$\;
	
    If a neuron exists in both $\mathbf{L}_i'^A$ and $\mathbf{L}_i'^B$, remove the neuron from them both.
    
    }
    \caption{Regroup algorithm.}
	\label{Algo}
\end{algorithm}

Based on Theorem \ref{theorem:approx}, we propose an algorithm to regroup the neurons such that conditions \eqref{eqt:conditions} are approximately met. 
It constructs the largest possible set $\mathbf{L}'^A_i$ and $\mathbf{L}'^B_i$ from all the neurons in $\mathbf{L}_i^A$ and $\mathbf{L}_i^B$ while $I(\mathbf{L}_i'^A;\mathbf{Y}^B)$ and $I(\mathbf{L}_i'^B;\mathbf{Y}^A)$ remain close to zero, such that $I(\mathbf{L}_i'^A;\mathbf{Y}^A)$ and $I(\mathbf{L}_i'^B;\mathbf{Y}^B)$ are approximately maximised.
To estimate $I(\mathbf{L}_i'^A;\mathbf{Y}^B)$ and $I(\mathbf{L}_i'^B;\mathbf{Y}^A)$, we use a  Kullback–Leibler-based mutual information upper bound estimator from~\cite{bib:Entropy17:Kolchinsky}.

Algorithm~\ref{Algo} illustrates the pseudocode to regroup the neurons such that the conditions in Theorem~\ref{theorem:PAM} are approximated met.
Central in Algorithm~\ref{Algo} is a greedy search in Lines 5-8 and 10-13.
In Lines 5-8, we search for the largest possible set of neuron $\mathbf{L}'^A_i$ while $I(\mathbf{L}'^A_i;\mathbf{Y}^B)$ remains approximately zero (smaller than a pre-defined threshold $\alpha$), such that $I(\mathbf{L}'^A_i;\mathbf{Y}^A)$ is approximately maximised. 
Similarly, in Lines 10-13, we approximately maximise $I(\mathbf{L}'^B_i;\mathbf{Y}^B)$ while keeping $I(\mathbf{L}'^B_i;\mathbf{Y}^A)$ close to zero.
According to Theorem~\ref{theorem:approx}, the conditions in Theorem~\ref{theorem:PAM} are approximately met.

\fakeparagraph{Practical Issue: How to Estimate Mutual Information}
We use a  Kullback–Leibler-based mutual information upper bound estimator from~\cite{bib:Entropy17:Kolchinsky} to estimate the upper bounds of $I(\mathbf{L}'^A_i;\mathbf{Y}^B)$ and $I(\mathbf{L}'^B_i;\mathbf{Y}^A)$.
Since the upper bounds are approximate, it is impossible to request them to be exactly zero.
Hence, we use a threshold parameter $\alpha$ to keep $I(\mathbf{L}'^A_i;\mathbf{Y}^B)$ and $I(\mathbf{L}'^B_i;\mathbf{Y}^A)$ close to zero.

\fakeparagraph{Practical Issue: How to Tune Threshold $\alpha$}
The parameter $\alpha$ affects the performance of ``\sysname \& prune''.
A larger $\alpha$ results in more neurons in $\mathbf{L}'^A_i$ and $\mathbf{L}'^B_i$ and fewer shared neurons in $\mathbf{L}'^{A,B}_i$.
In this case, the multitask network after ``\sysname \& prune'' performs worse in terms of efficiency when both tasks are executed concurrently, but better when only one task is executed (similar to ``baseline 1 \& prune'').
Conversely, a smaller $\alpha$ results in more shared neurons.
In this case, the multitask network after ``\sysname \& prune'' performs worse when only one task is executed, but better when both tasks are executed concurrently, (similar to ``baseline 2 \& prune'').

The parameter $\alpha$ can be empirically tuned as follows: 
\begin{enumerate}
    \item 
    Execute Algorithm~\ref{Algo} with a small $\alpha$.
    \item 
    Increase the value of $\alpha$ slightly and rerun Algorithm~\ref{Algo}.
    Since Lines 5-8 and 10-13 are greedy search, the results for the smaller $\alpha$ in Step 1 (\ie the already constructed neuron sets $\mathbf{L}'^A_i$ and $\mathbf{L}'^B_i$ ) can be reused, instead of starting with empty sets as in Line 4 and 9.
    \item 
    Iterate Step 2 till a satisfying balance among task combinations. 
    In each iteration of Step 2, we can reuse the neuron sets $\mathbf{L}'^A_i$ and $\mathbf{L}'^B_i$ from the last iteration.
\end{enumerate}
The impact of $\alpha$ is shown in \appref{appendix:alpha}.

\subsection{Extensions to ResNets}
\label{subsec:ResNets}

\begin{figure}[t]
\centering
\includegraphics[width=0.35\textwidth]{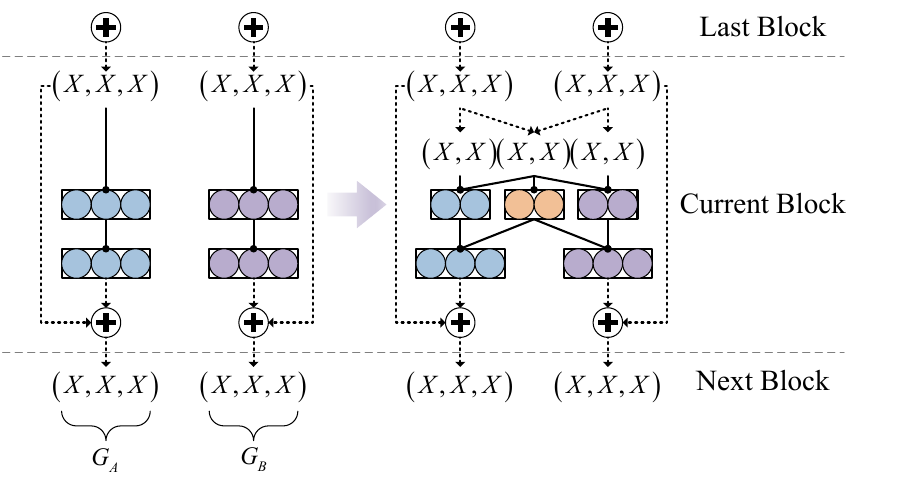}
\caption{Applying \sysname on residual blocks. Vectors are denoted as $(X,\cdots,X)$. Dotted lines are identical connections, and firm lines represent weighted connections for neurons.}
\label{fig:resnet}
\end{figure}

In order to support merging Residual Networks~\cite{bib:CVPR16:He}, \sysname needs to be slightly modified. 
As illustrated in \figref{fig:resnet}, the regrouping of the last layer in each residual block happens not directly after the weighted summation, but after the superposition with the shortcut connection and just before the vector is passed as inputs to the first layer in the next block.
This input vector of the first layer in each block is also regrouped using Algorithm~\ref{Algo} and then pruned at a later stage. 
This special treatment for the last layer in each residual block is consistent with ResNet compatible pruning methods such as~\cite{bib:CVPR19:Molchanov}, which can also prune the block outputs just before it is fed into the first layer in the next block.

\subsection{Extension to Three or More Tasks}
\label{appendix:extension}
When there are $K \geq 3$ tasks, we define the set of all the task as $\upsilon=\{t_1, \cdots, t_K\}$. 
The merged multitask network can be divided into subgraphs $\widetilde{G}_\tau$, where $\tau \subseteq \upsilon$ and $\tau \neq \emptyset$ is a nonempty subset of tasks.
Each vertex in $\widetilde{G}_\tau$ has paths to all the outputs $\widehat{\mathbf{Y}}^t$ with $t\in\tau$.
When a task combination (\ie a subset of tasks) $\tau$ is executed, only subgraph $\widetilde{G}_\tau$ is activated.
Layers in $\widetilde{G}_\tau$ is denoted as $\widetilde{\mathbf{L}}_i^\tau$.
The output layer for task combination $\tau$ is denoted as $\widehat{\mathbf{Y}}^\tau = \bigcup_{t \in \tau}\widehat{\mathbf{Y}}^t$, which is the prediction of ground-truth labels $\mathbf{Y}^\tau = \bigcup_{t \in \tau}\mathbf{Y}^t$.

\fakeparagraph{Extension of Theorem~\ref{theorem:PAM}}
For any pair of non-overlapped nonempty subsets of task $\tau_A$ and $\tau_B$ ($\tau_A \cap \tau_B = \emptyset$), define:
\begin{align}
    \mathbf{A}_i &= \widetilde{\mathbf{L}}_i^{\tau_A} \setminus \widetilde{\mathbf{L}}_i^{\tau_B} \\
    \mathbf{B}_i &= \widetilde{\mathbf{L}}_i^{\tau_B} \setminus \widetilde{\mathbf{L}}_i^{\tau_A} \\
    \mathbf{M}_i &= \widetilde{\mathbf{L}}_i^{\tau_A} \cap \widetilde{\mathbf{L}}_i^{\tau_B}
\end{align}

Then Theorem~\ref{theorem:PAM} is extended into:
\begin{theorem}
    \label{theorem:PAM_extended}
     If for all $i = 1,\cdots, N$ with $N = \min_{t\in \upsilon} N_t$, and for any pair of non-overlapped nonempty subsets of task $\tau_A$ and $\tau_B$, the following conditions are satisfied:
     \begin{equation}
        \begin{aligned}
         I(\mathbf{A}_i;\mathbf{B}_i;\mathbf{Y}^{\tau_A};\mathbf{Y}^{\tau_B}) = 0 \\ I(\mathbf{M}_i;\mathbf{Y}^{\tau_A}|\mathbf{A}_i,\mathbf{Y}^{\tau_B}) = 0 \\ I(\mathbf{M}_i;\mathbf{Y}^{\tau_B}|\mathbf{B}_i,\mathbf{Y}^{\tau_A}) = 0
         \end{aligned}
     \end{equation}
    then the computation cost of executing all task combinations can be minimised by the following $K$ non-conflicting optimisation problems that can be solved independently:
    \begin{equation}
    \text{For every $t \in \upsilon$: minimise } \medop{\sum_{i=1}^{N+1}}  \mathcal{R}_t(\widetilde{\mathbf{L}}_i^t) - \tilde{\xi}^t_i \cdot I(\widetilde{\mathbf{L}}_i^t;\mathbf{Y}^t)
    \end{equation}
\end{theorem}
Theorem~\ref{theorem:PAM_extended} can be proven by recursively applying Theorem~\ref{theorem:PAM}.

\fakeparagraph{Extension of \sysname}
The neuron sets $\mathbf{L}_i'^A$, $\mathbf{L}_i'^B$ and $\mathbf{L}_i'^{A,B}$ are extended to:
\begin{equation}
    \mathbf{L}_i'^\tau = \bigcap_{t \in \tau} \widetilde{\mathbf{L}}_i^t \setminus \bigcup_{t \notin \tau} \widetilde{\mathbf{L}}_i^t
\end{equation}

Note that neurons in $\mathbf{L}_i'^\tau$ are activated iff any task $t\in\tau$ is executed.
Now Algorithm~\ref{Algo} is extended to Algorithm~\ref{Algo_extended}.
And at step 5 of the \sysname workflow in \secref{subsec:framework}, we connect $\mathbf{L}_{i-1}'^{\tau_1} \to \mathbf{L}_i'^{\tau_2}$ iff $\tau_2 \subseteq \tau_1$. 

\begin{algorithm}[t]
	\SetAlgoLined
	\KwIn{$\mathbf{X}$, $\alpha$, $\mathbf{L}_i^t$, and $\mathbf{Y}^t$ for all $t\in \upsilon$}
	\KwOut{$\mathbf{L}_i'^\tau$ for all $\tau \subseteq \upsilon$ and $\tau \neq \emptyset$}
	\BlankLine
	$N \gets \min_{t\in \upsilon} N_t$\;
	
	$K \gets |\upsilon|$\;
	
	\For{$i\gets 1$ \KwTo $N$ }{
	    $\mathbf{S} \gets \bigcup_{t \in \upsilon} \mathbf{L}^t_i$\;
	    
	    \For {$n \gets 1$ \KwTo $K-1$}{
	        \For {any $\tau$ with $|\tau|=n$}{
        		$\mathbf{F} \gets \mathbf{S}$\; 
        		
        		$\mathbf{L}_i'^\tau \leftarrow \emptyset $\;
        		
        		$\mathbf{Y}^{\notin \tau} \gets \bigcup_{t \notin \tau}\mathbf{Y}^t$
        
        		\While{$I(\mathbf{L}_i'^\tau;\mathbf{Y}^{\notin \tau}) \leq \alpha$}{
        			$L_{i,\cdot} \gets \argmin_{L_{i,j} \in \mathbf{F}} I( \{L_{i,j}\}\cup \mathbf{L}_i'^\tau; \mathbf{Y}^{\notin \tau})$
        			
        			move the neuron $L_{i,\cdot}$ from  $\mathbf{F}$ to $\mathbf{L}_i'^\tau$
        			
        		}
    		}
    		Remove all selected neurons from $\mathbf{S}$: $\mathbf{S} \gets \mathbf{S} \setminus \bigcup_{|\tau| = n} \mathbf{L}'^\tau_i$
    		
    		Among all $\mathbf{L}_i'^\tau$, if a neuron exists in more than one set, remove the neuron from them all
    		
		}
    $\mathbf{L}_i'^\upsilon \gets \mathbf{S}$
    }
	\caption{Extending Algorithm~\ref{Algo} to over two tasks}
	\label{Algo_extended}
\end{algorithm}

It is worth mentioning that when tasks are highly related, the numbers of neurons in $\mathbf{L}_i^\tau$ with $1<|\tau|<K$ can be extremely small (as in our experiment on the LFW dataset in \appref{appendix:dataset}).
Therefore we can simplify Algorithm~\ref{Algo_extended} by fixing $n = 1$ and skip the remaining loops.
Every layer in the multitask network merged by the simplified \sysname contains only neuron sets $\mathbf{L}_i^t$ with $t\in\upsilon$ and one shared neuron set $\mathbf{L}_i^\upsilon$.
Shared neurons in $\mathbf{L}_i^\upsilon$ are always activated, while non-shared neurons in $\mathbf{L}_i^t$ are activated iff task $t$ is executed.
\section{Experiments}
\label{sec:experiments}
We compare different network merging schemes on whether lower computation is achieved when performing \textit{any subset} of tasks.

\subsection{Experiment Settings}
\label{subsec:experiment_settings}

\fakeparagraph{Baselines for Network Merging}
We compare \sysname with two merging schemes.
\begin{itemize}
    \item \textbf{Baseline 1}.
    It simply skips network merging in the ``merge \& prune'' framework.
    Therefore, no multitask network is constructed.
    As mentioned in \secref{sec:introduction}, this scheme optimises the pruning of \textit{single-task} networks.
    \item \textbf{Baseline 2}.
    Pre-trained single-task networks are merged as a multitask network by MTZ~\cite{bib:NIPS18:He}, a state-of-the-art network merging scheme.
    Applying MTZ in ``merge \& prune'' can minimise the computation cost of a multitask network when \textit{all tasks} are executed.
\end{itemize}

\fakeparagraph{Methods for Network Pruning}
Since we aim to compare different network merging schemes in the ``merge \& prune'' framework, we apply the same network pruning method on the neural network(s) constructed by different merging schemes.
To show that \sysname works with different pruning methods, we choose two state-of-the-art structured network pruning methods: one \cite{bib:ICML18:Dai} uses information theory based metrics (denoted as P1), and the other \cite{bib:CVPR19:Molchanov} uses sensitivity based metrics (denoted as P2).

The pruning methods are applied to the neural network(s) constructed by different merging schemes as follows.
For Baseline 1, each single-task network is pruned independently.
For the multitask network constructed with Baseline 2 and \sysname, we prune every subgraph for each individual task in an alternating manner (\eg task $A \to B \to C \to A \to B \to \cdots$) in order to balance between tasks.
However, only P2 is originally designed to prune a ResNet.
Hence we only experiment ResNets with P2.

\fakeparagraph{Datasets and Single-Task Networks}
We define tasks from three datasets: Fashion-MNIST~\cite{bib:arXiv17:Xiao}, CelebA~\cite{bib:ICCV15:Liu}, and LFW~\cite{bib:NIPS12:Huang}.
Fashion-MNIST and CelebA each contains \textit{two} tasks.
LFW contains \textit{five} tasks.
We use LeNet-5~\cite{bib:PIEEE98:LeCun} as pre-trained single-task networks for tasks derived from Fashion-MNIST, and VGG-16~\cite{bib:arXiv14:Simonyan} for tasks from CelebA and LFW.
We also use ResNet-18 and ResNet-34~\cite{bib:CVPR16:He} as pre-trained single-task networks for CelebA.
See \appref{appendix:dataset} for more details of dataset setup and the inference accuracy and FLOPs of the \textit{pre-trained} single-task networks.

\fakeparagraph{Evaluation Metrics}
For a given set of tasks, we aim to minimise the computation cost of all task combinations.
To assess computation cost independent of hardware, we use the number of floating point operations (FLOP) as the metric.
For fair comparison, the network(s) constructed by different merging schemes are pruned while preserving almost the same inference accuracy.
To quantify the performance advantage of \sysname \textit{over baselines} over all task combinations, we adopt the following two single-valued criteria:
\begin{itemize}
    \item \textbf{Average Gain}.
    This metric measures the averaged computation cost reduction of ``\sysname \& prune'' over ``baseline \& prune'' across \textit{all task combinations}. 
    For example, given two tasks $A$ and $B$, there are three task combinations: $A$, $B$ and $A\&B$. 
    When executing these task combinations, the FLOPs of the network after ``\sysname \& prune'' are $c_A^P$, $c_B^P$ and $c_{A,B}^P$, respectively. 
    After ``baseline 1 \& prune'', the FLOPs are $c_A^{B1}$, $c_B^{B1}$ and $c_{A,B}^{B1}$, respectively.
    The average gain over baseline 1 is calculated as $\frac{1}{3} (c_A^{B1}/c_A^P + c_B^{B1}/c_B^P + c_{A,B}^{B1}/c_{A,B}^P)$.
    \item \textbf{Peak Gain}.
    This metric measures the maximal computation cost reduction across \textit{all task combinations}.
    Using the same example and notations as above, the peak gain over baseline 1 is calculated as $\max \{c_A^{B1}/c_A^P, c_B^{B1}/c_B^P, c_{A,B}^{B1}/c_{A,B}^P\}$.
\end{itemize}

All experiments are implemented with TensorFlow and conducted on a workstation with Nvidia RTX 2080 Ti GPU.

\subsection{Main Experiment Results}

\begin{figure*}[t]
\centering
     \subfloat[LeNet/Fashion-MNIST]{
     	\includegraphics[width=0.22\textwidth]{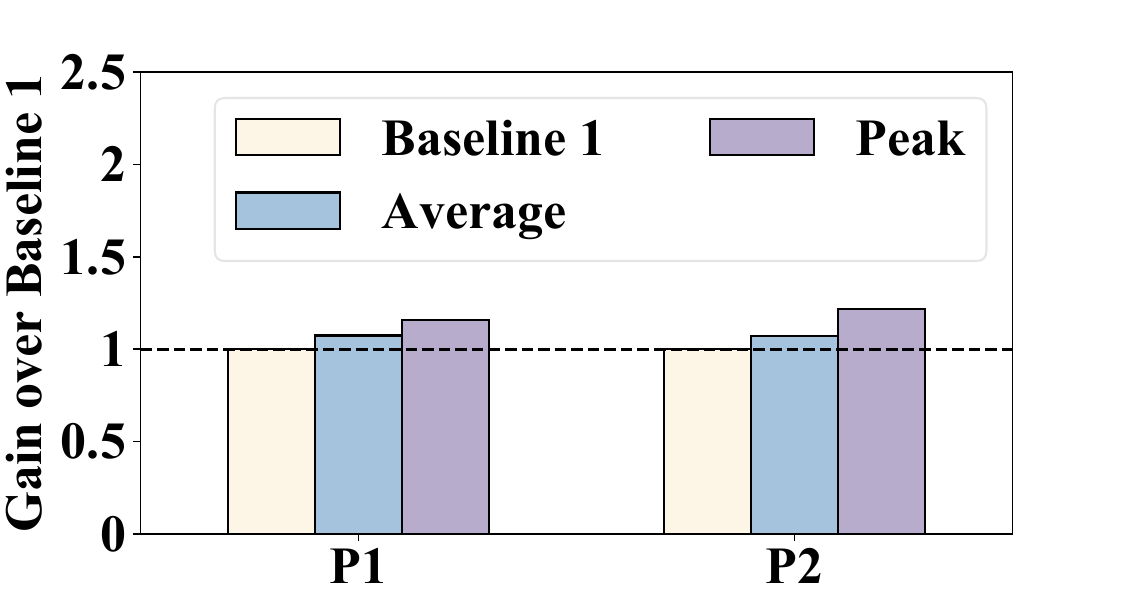}}\label{fig:mnist_b1}
     \subfloat[VGG/CelebA]{
     	\includegraphics[width=0.22\textwidth]{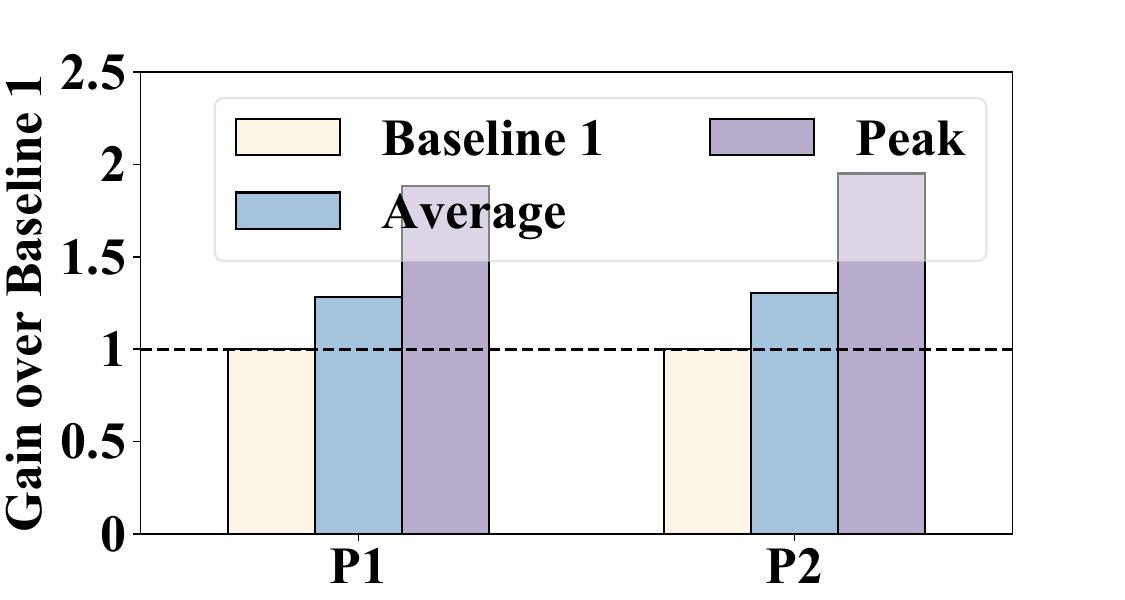}}\label{fig:celeba_b1}
     \subfloat[VGG/LFW]{
     	\includegraphics[width=0.22\textwidth]{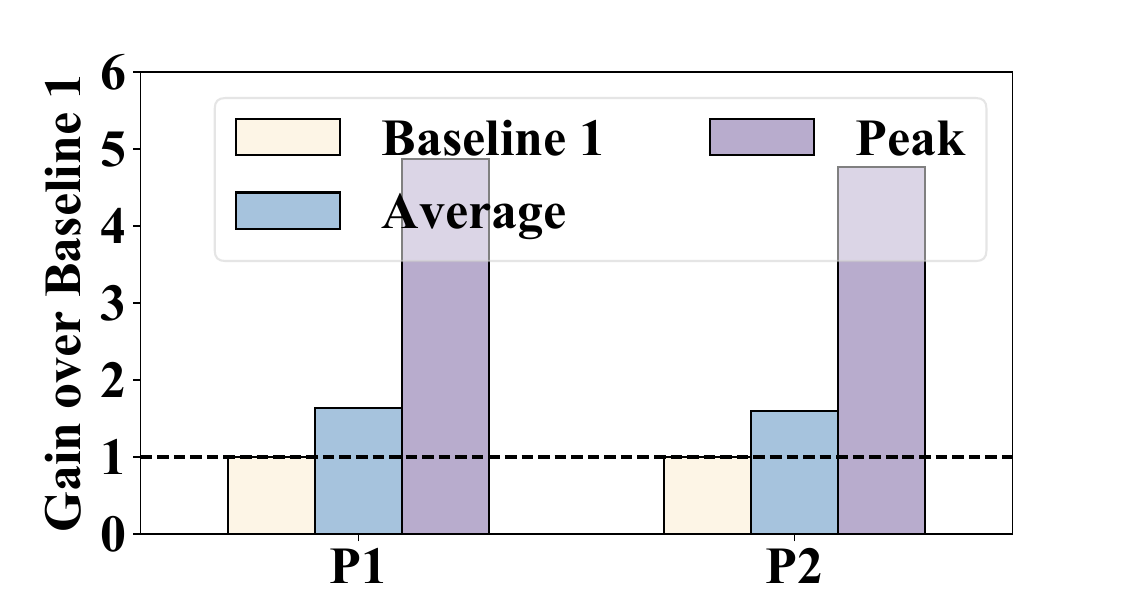}}\label{fig:flw_b1}
     \subfloat[ResNet/CelebA]{
     	\includegraphics[width=0.22\textwidth]{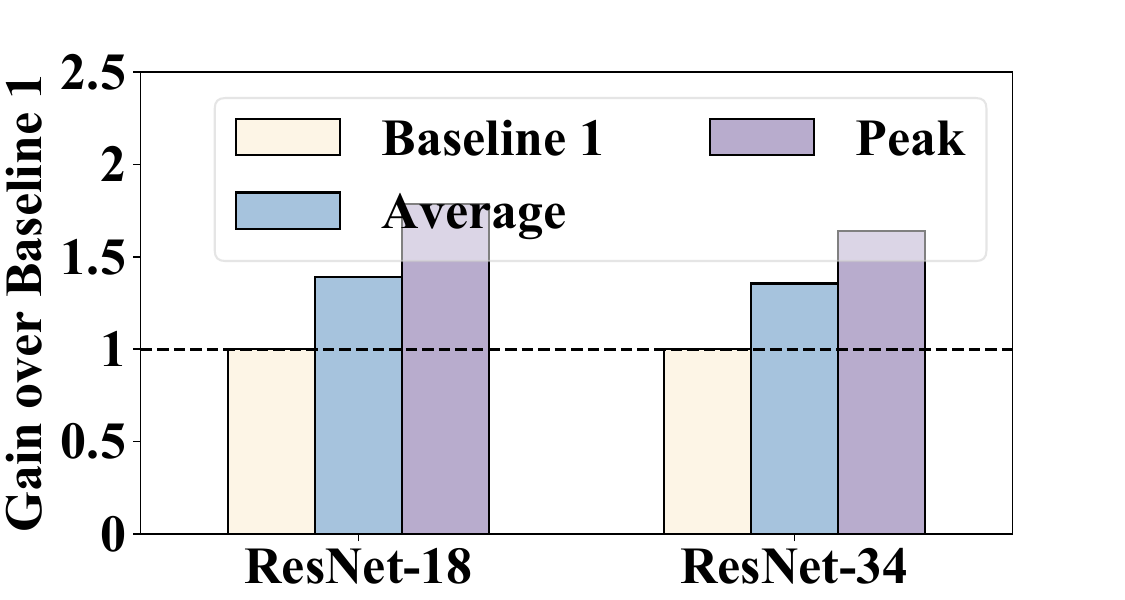}}\label{fig:resnet_b1}
     \\
     \subfloat[LeNet/Fashion-MNIST]{
     	\includegraphics[width=0.22\textwidth]{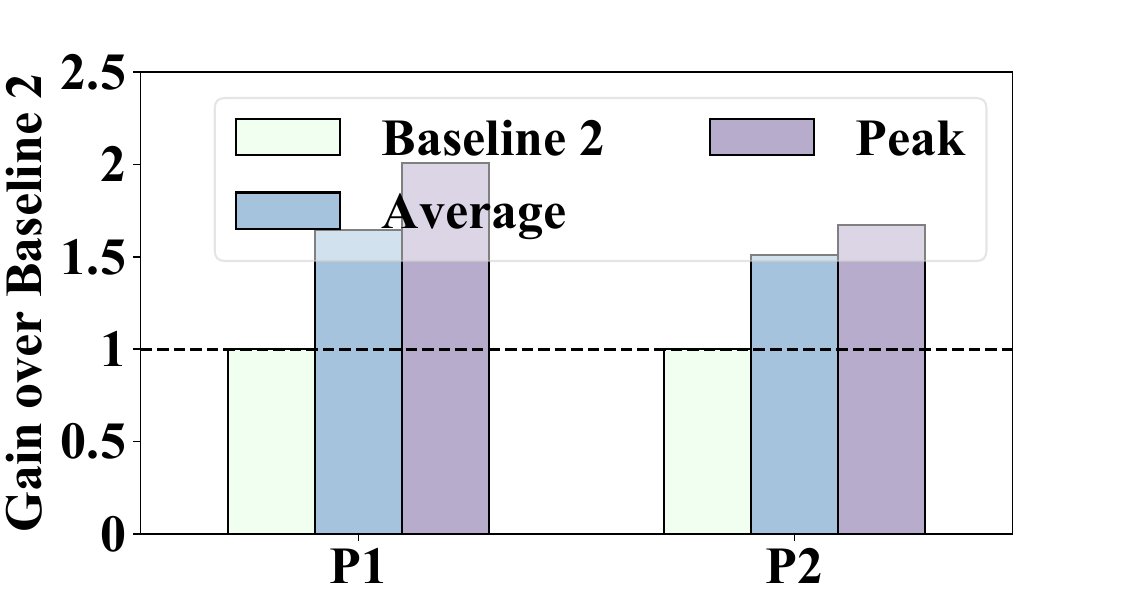}}\label{fig:mnist_b2}
     \subfloat[VGG/CelebA]{
     	\includegraphics[width=0.22\textwidth]{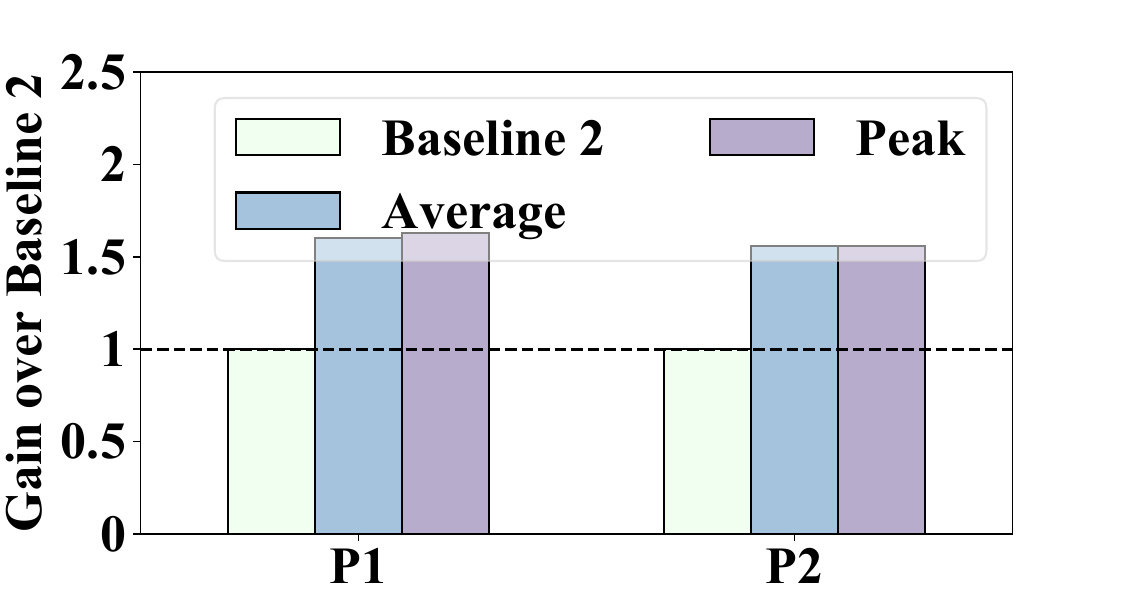}}\label{fig:celeba_b2}
     \subfloat[VGG/LFW]{
     	\includegraphics[width=0.22\textwidth]{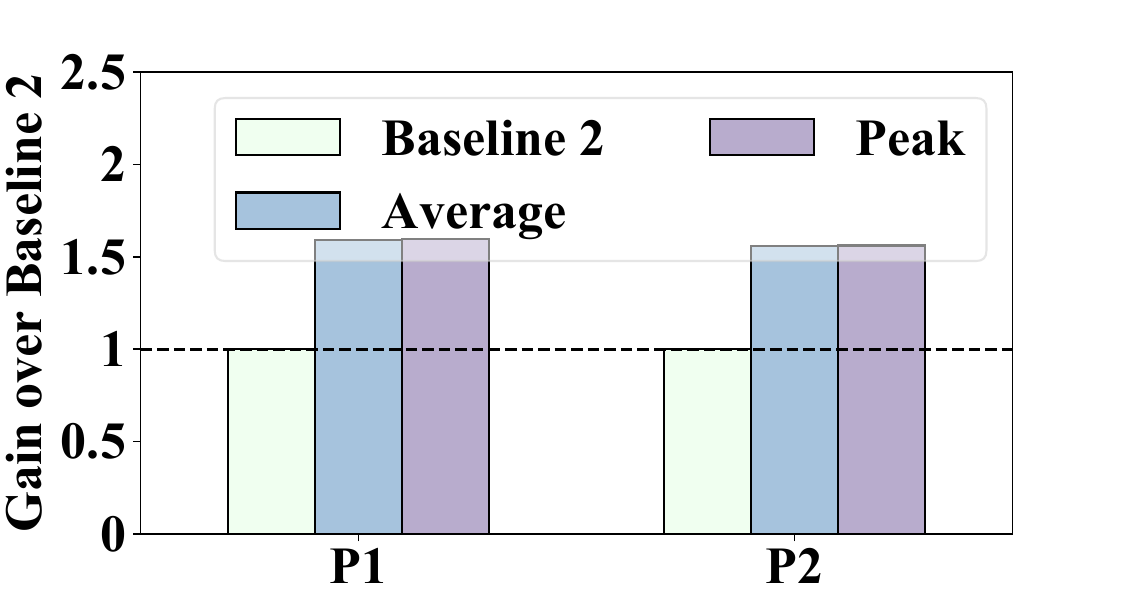}}\label{fig:flw_b2}
     \subfloat[ResNet/CelebA]{
     	\includegraphics[width=0.22\textwidth]{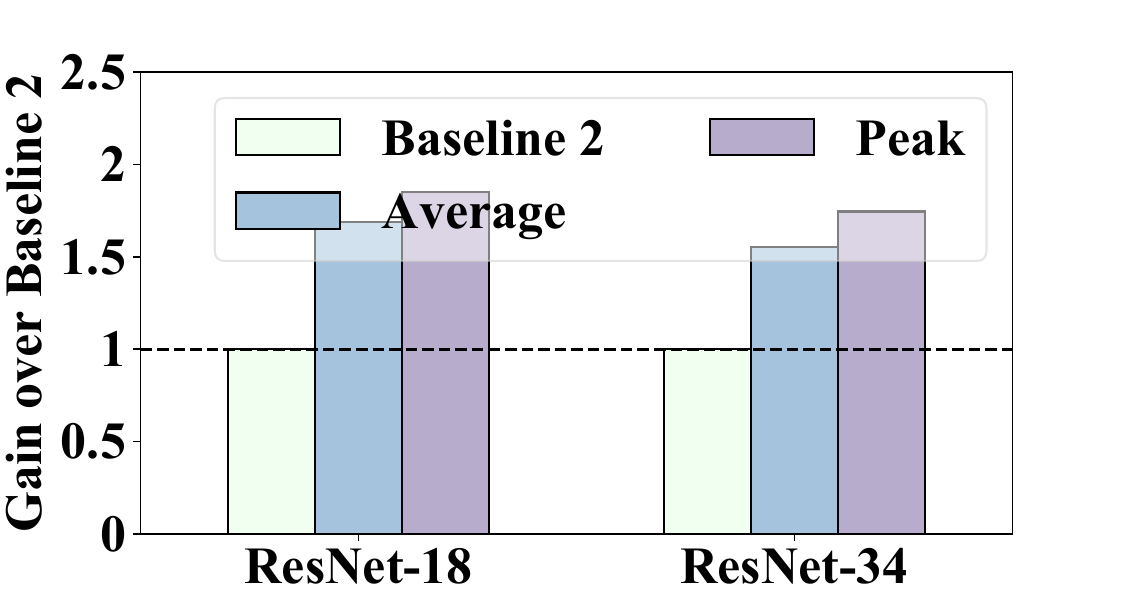}}\label{fig:resnet_b2} 
     \caption{Average and peak gain of \sysname over baselines in different combinations of models, datasets, and pruning methods. The upper row (a)-(d) shows the gain of \sysname over baseline 1. The lower row (e)-(h) shows the gain of \sysname over baseline 2. Note that the average and peak gain of each baseline is $1$ by definition.}
     \label{fig:main_results}
\end{figure*}

\fakeparagraph{Overall Performance Gain}
\figref{fig:main_results} shows the average and peak gains of \sysname over the two baselines tested with different models (LeNet-5, VGG-16, ResNet-18, RestNet-34), datasets (Fashion-MNIST, CelebA, LFW), and pruning methods (P1, P2).
The detailed FLOPs and inference accuracy on task merging (Fashion-MNIST and CelebA) are listed in \tabref{tab:fashion_mnist}, \tabref{tab:celeba}, \tabref{tab:lfw} and \tabref{tab:resnet}.

Compared with baseline 1, \sysname achieves $1.07\times$ to $1.64\times$ average gain and $1.16\times$ to $4.87\times$ peak gain.
Compared with baseline 2, \sysname achieves $1.51\times$ to $1.69\times$ average gain and $1.56\times$ to $2.01\times$ peak gain.
In general, \sysname has significant performance advantage over both baselines across datasets and network architectures.

\begin{table}[ht]
    \caption{Test accuracy and computation cost of all tasks combinations with LeNet-5 on Fashion-MNIST pruned by P1/P2.}
    \label{tab:fashion_mnist}
    \centering
    \scriptsize
    \begin{tabular}{llllllll}
        \toprule
        \multirow{2}{*}{Pruning} & \multirow{2}{*}{Tasks} & \multicolumn{3}{c}{Accuracy} & \multicolumn{3}{c}{FLOPs ($\times10^6$)}\\
       \cmidrule(lr){3-5} \cmidrule(lr){6-8} 
        & &  B1 & B2 & \sysname & B1 & B2 & \sysname \\
        \midrule
		\multirow{3}{*}{P1} & A	    &   95.42\%	&	95.30\%	&	94.67\%	&	28.34	&	52.58   &	28.49	\\
		& B	    &	96.30\%	&	96.40\%	&	95.70\%	&	28.34	&	52.58   &	26.16	\\ 
		& A\&B    &	95.86\%	&	95.85\%	&	95.19\%	&	56.69	&	52.58   &	48.68	\\
        \midrule
        \multirow{3}{*}{P2} & A	    &   95.82\%	&	95.73\%	&	95.70\%	&	18.64	&	31.19   &	18.65	\\
		& B	    &	96.46\%	&	96.72\%	&	96.38\%	&	18.64	&	31.19   &	18.65	\\ 
		& A\&B    &	96.14\%	&	96.22\%	&	96.04\%	&	37.27	&	31.19   &	26.48	\\
        \bottomrule
    \end{tabular}
\end{table}

\begin{table}[ht]
    \caption{Test accuracy and computation cost of all tasks combinations with VGG-16 on CelebA pruned by P1/P2.}
    \label{tab:celeba}
    \centering
    \scriptsize
    \begin{tabular}{lllllllll}
        \toprule
        \multirow{2}{*}{Pruning} & \multirow{2}{*}{Tasks} & \multicolumn{3}{c}{Accuracy} & \multicolumn{3}{c}{FLOPs ($\times10^6$)}\\
       \cmidrule(lr){3-5} \cmidrule(lr){6-8} 
         &  &  B1 & B2 & \sysname & B1 & B2 & \sysname \\
        \midrule
		\multirow{3}{*}{P1} & A	    &   89.45\%	&	89.09\%	&	89.60\%	&	4.52	&	7.3   &	4.48	\\
		& B	    &	87.81\%	&	87.69\%	&	88.00\%	&	4.32	&	7.3   &	4.49	\\ 
		& A\&B    &	88.63\%	&	88.39\%	&	88.80\%	&	8.85	&	7.3   &	4.70	\\
        \midrule
		\multirow{3}{*}{P2} & A	    &   90.34\%	&	90.27\%	&	90.36\%	&	153.13	&	243.20   &	155.82	\\
		& B	    &	88.84\%	&	88.74\%	&	88.76\%	&	152.65	&	243.20   &	155.84	\\ 
		& A\&B    &	89.59\%	&	89.51\%	&	89.56\%	&	305.78	&	243.20   &	156.74	\\
        \bottomrule
    \end{tabular}
\end{table}

\begin{table}[ht]
     \caption{Test accuracy and computation cost of all tasks combinations with VGG-16 on LFW pruned by P1/P2.}
     \label{tab:lfw}
     \centering
     \scriptsize
     \begin{tabular}{llllllll}
         \toprule
         \multirow{2}{*}{Pruning} & \multirow{2}{*}{Tasks} & \multicolumn{3}{c}{Accuracy} & \multicolumn{3}{c}{FLOPs ($\times10^6$)}\\
       \cmidrule(lr){3-5} \cmidrule(lr){6-8} 
      & &  B1 & B2 & \sysname & B1 & B2 & \sysname \\
         \midrule
 		\multirow{30}{*}{P1} & A	&	89.77\%	&	89.49\%	&	89.87\%	&	7.96	&	12.66	&	7.94	\\
 		& B	&	82.81\%	&	82.82\%	&	82.14\%	&	7.91	&	12.66   &	7.95	\\
 		& C	&	83.20\%	&	82.68\%	&	83.30\%	&	7.94	&	12.66	&	7.94	\\
 		& D	&	85.74\%	&	86.45\%	&	86.03\%	&	7.58	&	12.66	&	7.93	\\
 		& E	&	87.10\%	&	86.52\%	&	86.90\%	&	7.87	&	12.66	&	7.93	\\  
 		\cmidrule(lr){2-8}
 		& A\&B	&	86.29\%	&	86.16\%	&	86.00\%	&	15.87	&	12.66	&	7.98	\\
 		& A\&C	&	86.48\%	&	86.09\%	&	86.59\%	&	15.90	&	12.66	&	7.97	\\
 		& A\&D	&	87.75\%	&	87.97\%	&	87.95\%	&	15.54	&	12.66	&	7.97	\\
 		& A\&E	&	88.44\%	&	88.01\%	&	88.39\%	&	15.84	&	12.66	&	7.96	\\
 		& B\&C	&	83.00\%	&	82.75\%	&	82.72\%	&	15.85	&	12.66	&	7.98	\\
 		& B\&D	&	84.28\%	&	84.64\%	&	84.09\%	&	15.49	&	12.66	&	7.97	\\
 		& B\&E	&	84.95\%	&	84.67\%	&	84.52\%	&	15.79	&	12.66	&	7.97	\\
 		& C\&D	&	84.47\%	&	84.57\%	&	84.66\%	&	15.52	&	12.66	&	7.96	\\
 		& C\&E	&	85.15\%	&	84.60\%	&	85.10\%	&	15.81	&	12.66	&	7.96	\\
 		& D\&E	&	86.42\%	&	86.49\%	&	86.47\%	&	15.45	&	12.66	&	7.96	\\  
 		\cmidrule(lr){2-8}
 		& A\&B\&C	&	85.26\%	&	85.00\%	&	85.10\%	&	23.81	&	12.66	&	8.01	\\
 		& A\&B\&D	&	86.11\%	&	86.25\%	&	86.01\%	&	23.45	&	12.66	&	8.01	\\
 		& A\&B\&E	&	86.56\%	&	86.28\%	&	86.30\%	&	23.75	&	12.66	&	8.00	\\
 		& A\&C\&D	&	86.24\%	&	86.21\%	&	86.40\%	&	23.48	&	12.66	&	8.00	\\
 		& A\&C\&E	&	86.69\%	&	86.23\%	&	86.69\%	&	23.78	&	12.66	&	7.99	\\
 		& A\&D\&E	&	87.54\%	&	87.49\%	&	87.60\%	&	23.42	&	12.66	&	7.99	\\
 		& B\&C\&D	&	83.92\%	&	83.98\%	&	83.82\%	&	23.43	&	12.66	&	8.01	\\
 		& B\&C\&E	&	84.37\%	&	84.01\%	&	84.11\%	&	23.73	&	12.66	&	8.00	\\
 		& B\&D\&E	&	85.22\%	&	85.26\%	&	85.02\%	&	23.37	&	12.66	&	8.00	\\
 		& C\&D\&E	&	85.35\%	&	85.22\%	&	85.41\%	&	23.39	&	12.66	&	7.99	\\  
 		\cmidrule(lr){2-8}
 		& A\&B\&C\&D	&	85.38\%	&	85.36\%	&	85.34\%	&	31.39	&	12.66	&	8.04	\\
 		& A\&B\&C\&E	&	85.72\%	&	85.38\%	&	85.55\%	&	31.69	&	12.66	&	8.03	\\
 	    & A\&B\&D\&E	&	86.35\%	&	86.32\%	&	86.23\%	&	31.33	&	12.66	&	8.03	\\
 		& A\&C\&D\&E	&	86.45\%	&	86.29\%	&	86.53\%	&	31.36	&	12.66	&	8.02	\\
 		& B\&C\&D\&E	&	84.71\%	&	84.62\%	&	84.59\%	&	31.31	&	12.66	&	8.03	\\  \cmidrule(lr){2-8}
 		& A\&B\&C\&D\&E	&	85.72\%	&	85.59\%	&	85.65\%	&	39.27	&	12.66	&	8.06	\\
         \midrule
 	\multirow{30}{*}{P2} &	A	&	89.57\%	&	89.38\%	&	89.24\%	&	22.91	&	36.33	&	23.28	\\
 	&	B	&	81.96\%	&	83.15\%	&	83.39\%	&	23.16	&	36.33	&	23.29	\\
 	&	C	&	82.96\%	&	81.61\%	&	82.10\%	&	22.93	&	36.33	&	23.28	\\
 	&	D	&	85.04\%	&	85.12\%	&	85.29\%	&	21.16	&	36.33	&	23.27	\\
 	&	E	&	86.43\%	&	85.81\%	&	85.57\%	&	21.29	&	36.33	&	23.27	\\  
 	\cmidrule(lr){2-8}
 	&	A\&B	&	85.76\%	&	86.27\%	&	86.31\%	&	46.07	&	36.33	&	23.32	\\
 	&	A\&C	&	86.26\%	&	85.50\%	&	85.67\%	&	45.84	&	36.33	&	23.31	\\
 	&	A\&D	&	87.31\%	&	87.25\%	&	87.27\%	&	44.07	&	36.33	&	23.30	\\
 	&	A\&E	&	88.00\%	&	87.60\%	&	87.41\%	&	44.20	&	36.33	&	23.30	\\
 	&	B\&C	&	82.46\%	&	82.38\%	&	82.75\%	&	46.08	&	36.33	&	23.31	\\
 	&	B\&D	&	83.50\%	&	84.14\%	&	84.34\%	&	44.31	&	36.33	&	23.31	\\
 	&	B\&E	&	84.19\%	&	84.48\%	&	84.48\%	&	44.45	&	36.33	&	23.31	\\
 	&	C\&D	&	84.00\%	&	83.37\%	&	83.69\%	&	44.09	&	36.33	&	23.30	\\
 	&	C\&E	&	84.69\%	&	83.71\%	&	83.83\%	&	44.22	&	36.33	&	23.30	\\
 	&	D\&E	&	85.74\%	&	84.47\%	&	85.43\%	&	42.45	&	36.33	&	23.29	\\  
 	\cmidrule(lr){2-8}
 	&	A\&B\&C	&	84.83\%	&	84.71\%	&	84.91\%	&	68.99	&	36.33	&	23.34	\\
 	&	A\&B\&D	&	85.52\%	&	85.88\%	&	85.97\%	&	67.22	&	36.33	&	23.34	\\
 	&	A\&B\&E	&	85.99\%	&	86.11\%	&	86.07\%	&	67.36	&	36.33	&	23.34	\\
 	&	A\&C\&D	&	85.86\%	&	85.37\%	&	85.54\%	&	67.00	&	36.33	&	23.33	\\
 	&	A\&C\&E	&	86.32\%	&	85.60\%	&	85.64\%	&	67.13	&	36.33	&	23.32	\\
 	&	A\&D\&E	&	87.01\%	&	86.77\%	&	86.70\%	&	65.36	&	36.33	&	23.32	\\
 	&	B\&C\&D	&	83.32\%	&	83.29\%	&	83.59\%	&	67.24	&	36.33	&	23.34	\\
 	&	B\&C\&E	&	83.78\%	&	83.52\%	&	83.69\%	&	67.37	&	36.33	&	23.33	\\
 	&	B\&D\&E	&	84.48\%	&	84.69\%	&	84.75\%	&	65.60	&	36.33	&	23.33	\\
 	&	C\&D\&E	&	84.81\%	&	84.18\%	&	84.32\%	&	65.38	&	36.33	&	23.32	\\  
 	\cmidrule(lr){2-8}
 	&	A\&B\&C\&D	&	84.88\%	&	84.82\%	&	85.00\%	&	90.15	&	36.33	&	23.37	\\
 	&	A\&B\&C\&E	&	85.23\%	&	84.99\%	&	85.07\%	&	90.28	&	36.33	&	23.36	\\
 	&	A\&B\&D\&E	&	85.75\%	&	85.87\%	&	85.87\%	&	88.51	&	36.33	&	23.36	\\
 	&	A\&C\&D\&E	&	86.00\%	&	85.48\%	&	85.55\%	&	88.29	&	36.33	&	23.35	\\
 	&	B\&C\&D\&E	&	84.10\%	&	83.92\%	&	84.09\%	&	88.53	&	36.33	&	23.36	\\  
 	\cmidrule(lr){2-8}
 	&	A\&B\&C\&D\&E	&	85.19\%	&	85.01\%	&	85.12\%	&	111.44	&	36.33	&	23.39	\\
     \bottomrule
     \end{tabular}
 \end{table}

\begin{table}[ht]
    \caption{Test accuracy and computation cost with ResNet-18/ResNet-34 on CelebA pruned by P1.}
    \label{tab:resnet}
    \centering
    \scriptsize
    \begin{tabular}{lllllllll}
        \toprule
        \multirow{2}{*}{Model} & \multirow{2}{*}{Tasks} & \multicolumn{3}{c}{Accuracy} & \multicolumn{3}{c}{FLOPs ($\times10^6$)}\\
       \cmidrule(lr){3-5} \cmidrule(lr){6-8} 
         &   &  B1 & B2 & \sysname & B1 & B2 & \sysname \\
        \midrule
		\multirow{3}{*}{ResNet-18} & A	    &   89.83\%	&	89.30\%	&	89.93\%	&	5.72	&	8.84   &	4.78	\\
		& B	    &	88.25\%	&	88.20\%	&	88.36\%	&	5.72	&	8.84   &	4.83	\\ 
		& A\&B    &	89.04\%	&	88.75\%	&	89.15\%	&	11.44	&	8.84   &	6.40	\\
       \midrule
		\multirow{3}{*}{ResNet-34} & A	    &   89.99\%	&	89.70\%	&	90.05\%	&	8.43	&	12.11   &	6.94	\\
		& B	    &	88.44\%	&	88.98\%	&	88.42\%	&	8.43	&	12.11   &	6.94	\\ 
		& A\&B    &	89.22\%	&	89.34\%	&	89.24\%	&	16.86	&	12.11   &	10.29	\\
        \bottomrule
    \end{tabular}
\end{table}

\begin{figure}[t]
\centering
     \subfloat[]{
     	\includegraphics[width=0.30\linewidth]{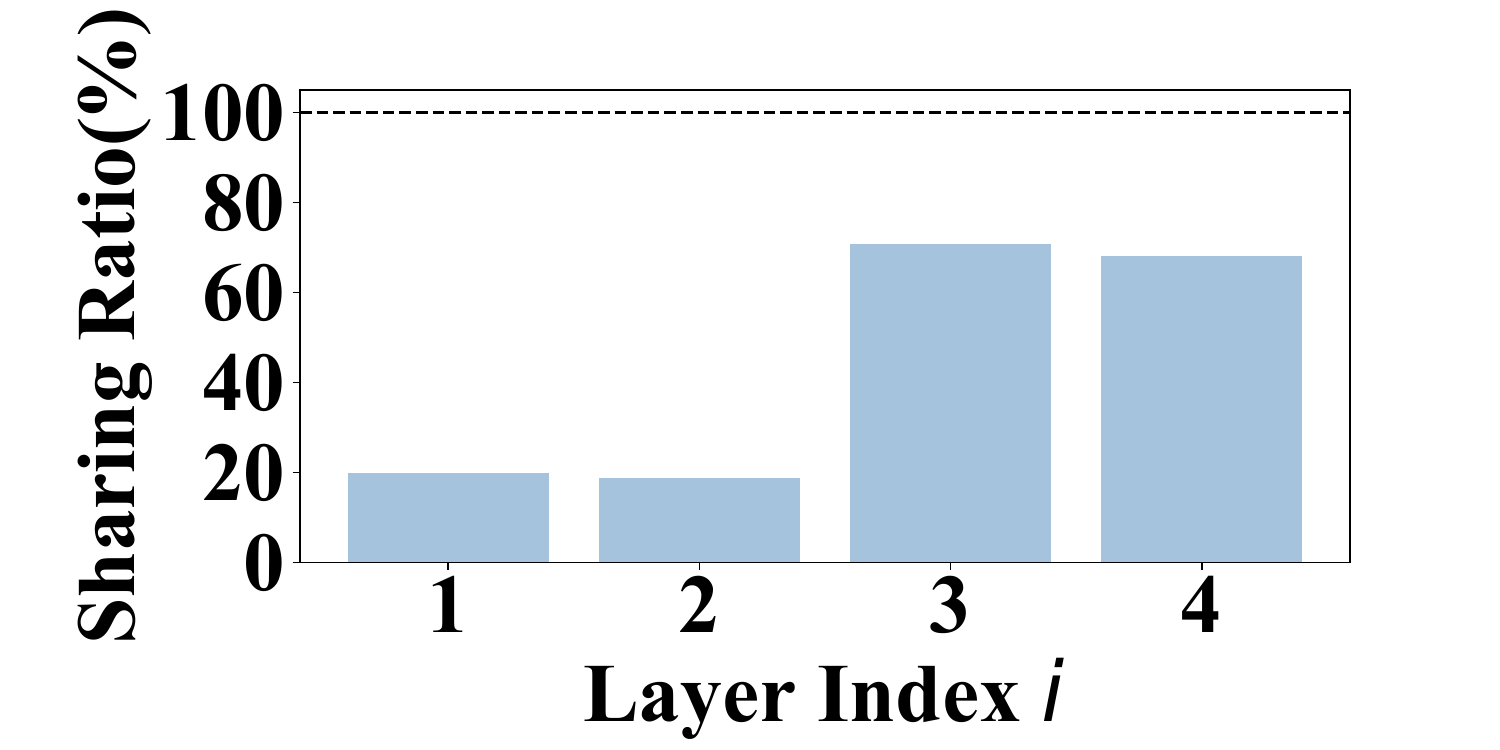}}
     \subfloat[]{
     	\includegraphics[width=0.30\linewidth]{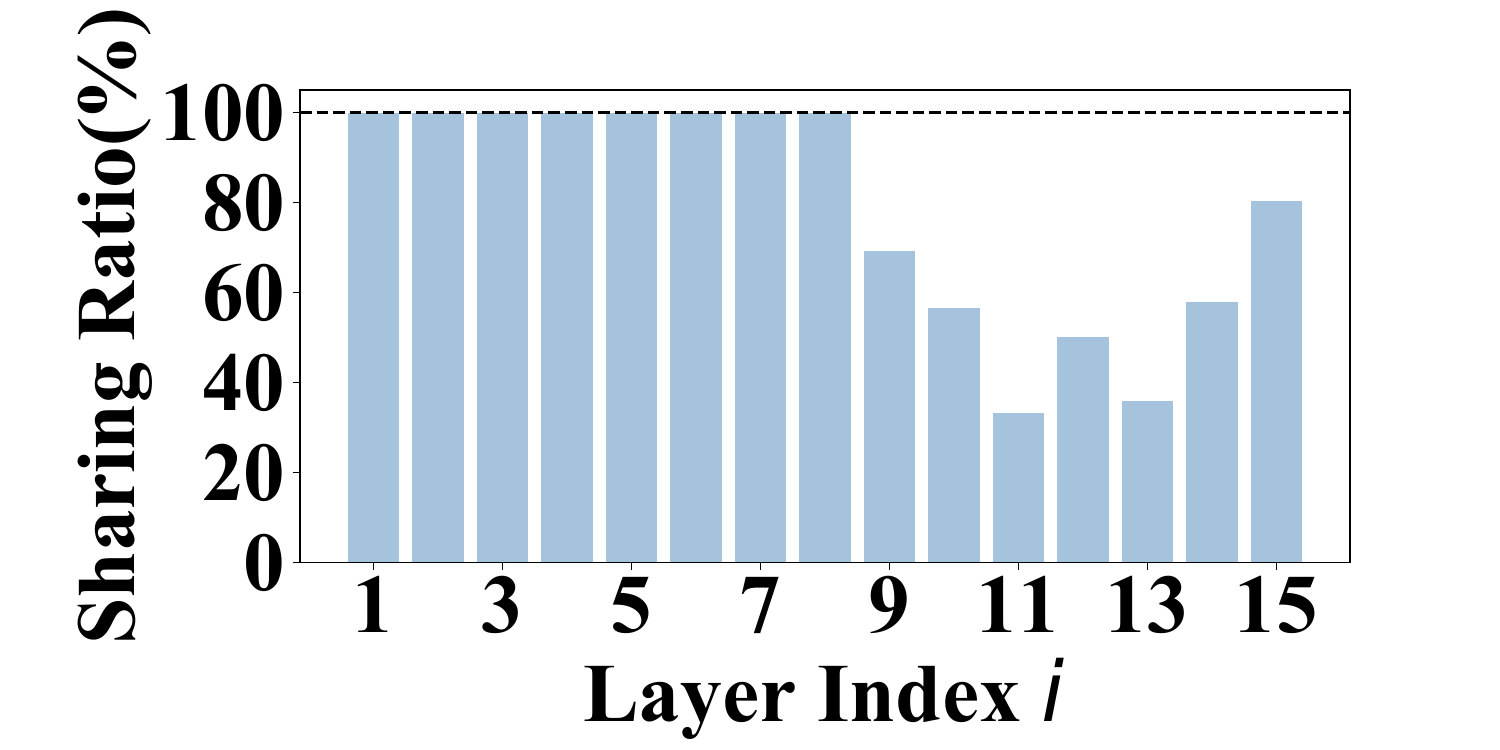}}
     \subfloat[]{
         \label{subfig:P1+LFW}
     	\includegraphics[width=0.30\linewidth]{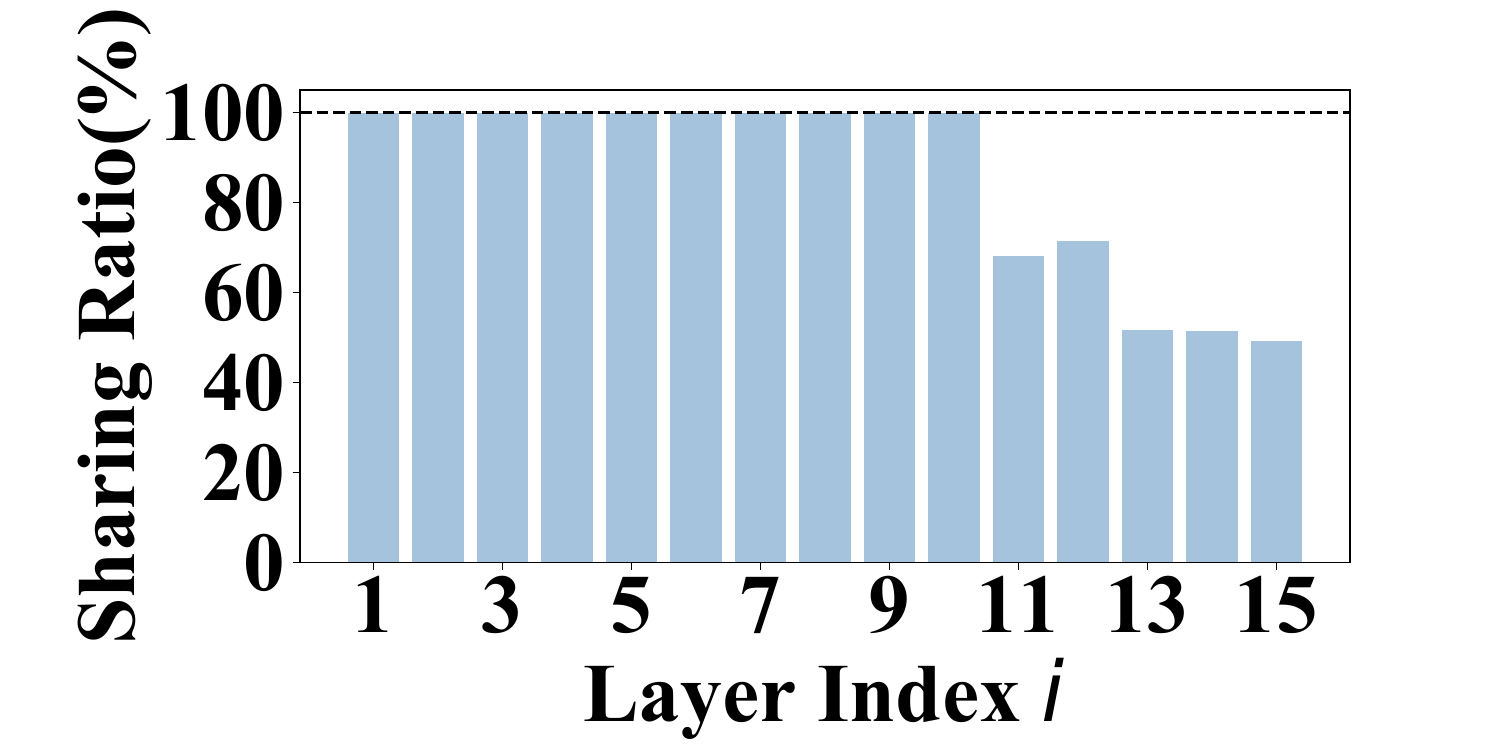}}\\
     \subfloat[]{
     	\includegraphics[width=0.30\linewidth]{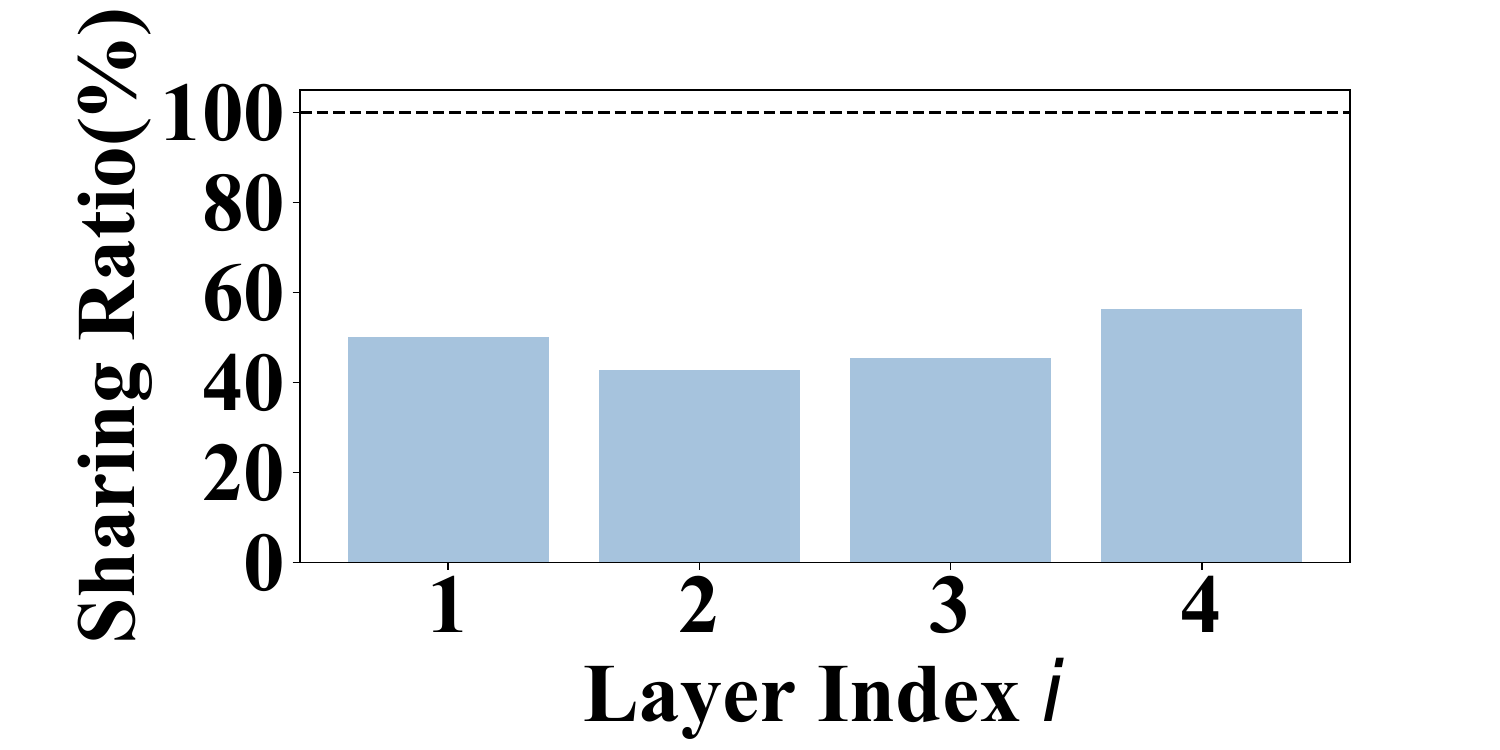}}
     \subfloat[]{
     	\includegraphics[width=0.30\linewidth]{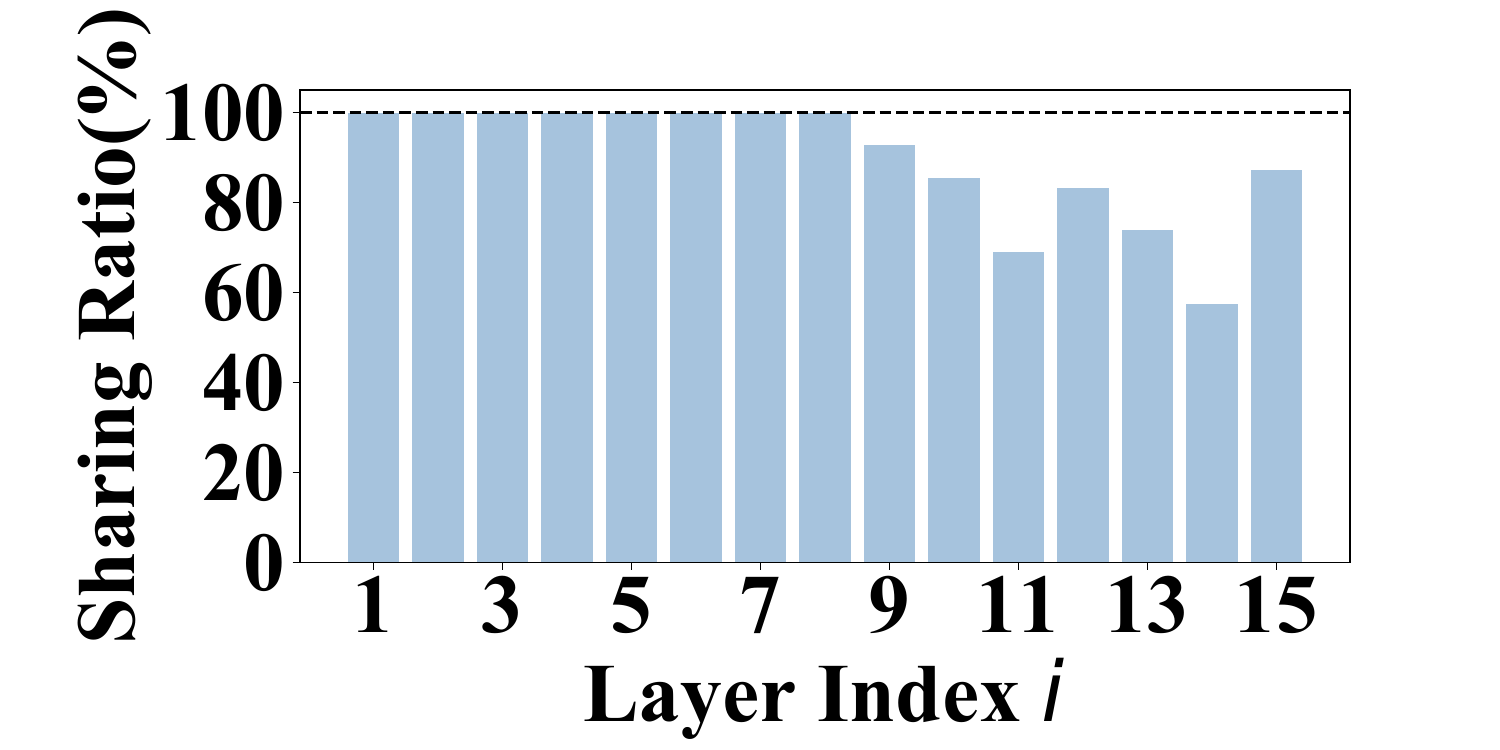}}
     \subfloat[]{
         \label{subfig:P2+LFW}
     	\includegraphics[width=0.30\linewidth]{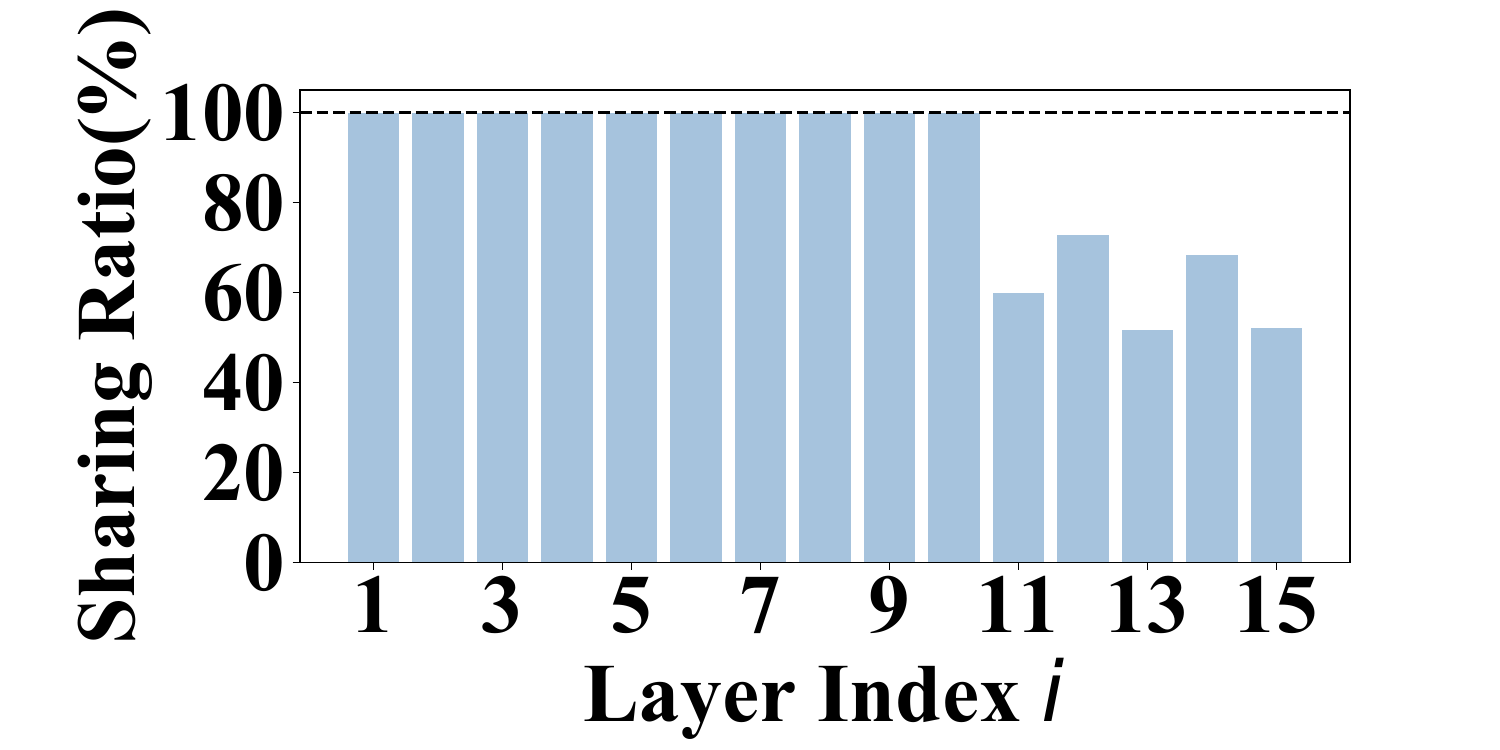}}
     \caption{Sharing ratio of each layer after ``\sysname \& prune (P1 or P2)'' on (a) LeNet/Fashion-MNIST with P1, (b) VGG/CelebA with P1, (c) VGG/LFW with P1, (d) LeNet/Fashion-MNIST with P2, (e) VGG/CelebA with P2, and (f) VGG/LFW with P2.
     In each layer, the sharing ratio is calculated as the number of shared neurons in $\mathbf{L}'^{A,B}_i$, divided by all neurons in $\mathbf{L}^{A,B}_i$.
     It ranges from $0\%$ to $100\%$.}
     \vspace{-3mm}
     \label{fig:sharing_ratio}
 \end{figure}

\fakeparagraph{Effectiveness of \sysname}
From \figref{fig:main_results}, the performance gain of \sysname varies across baselines and datasets.
Such variations in average and peak gains are influenced by \textit{how many neurons are shared} and \textit{how many networks are merged}.
\figref{fig:sharing_ratio} shows how many neurons (kernels) are shared after ``\sysname \& prune'' on LeNet-5 and VGG-16.
\begin{itemize}
    \item 
    \textbf{The more neurons shared, the higher gain \sysname has over baseline 1.}
    ``Baseline 1 \& prune'' can effectively reduce the computation cost when \textit{only one} task is performed. 
    However, when many neurons can be shared (see \figref{fig:sharing_ratio}(b), (c), (e), and (f)), baseline 1 is sub-optimal when multiple tasks are executed simultaneously, as it is unable to reduce computation by sharing neurons.
    This is why \sysname outperforms baseline 1 more on CelebA and LFW.
    \item
    \textbf{The fewer neurons shared, the higher gain \sysname has over baseline 2.}
    ``Baseline 2 \& prune'' can effectively reduce the computation cost via neuron sharing when \textit{all} tasks are performed simultaneously.
    However, when only few neurons can be shared (see \figref{fig:sharing_ratio}(a) and (d)), the multitask network merged by baseline 2 cannot shut down the unnecessary neurons when not all tasks are executed, and hence yields sub-optimal computation cost.
    This is why \sysname outperforms baseline 2 more on Fashion-MNIST.
    \item
    \textbf{The more networks merged, the higher gain \sysname has over both baselines.}
    As the number of single-task networks (tasks) increases, ``\sysname \& prune'' can either share more neurons and yield lower computation than ``baseline 1 \& prune'', or shut down more unnecessary neurons and yield lower computation than ``baseline 2 \& prune''.
    Therefore the performance gain of \sysname over baseline 1 on LFW is such significantly higher than on CelebA.
    This is also the reason why the performance gain of \sysname over baseline 2 on LFW is not much lower than on CelebA, although on LFW we have the highest degree of sharing.
\end{itemize}
 
\fakeparagraph{Takeaways}
Although the performance of \sysname varies across tasks, it achieves consistently solid advantages over both baselines. 
We may conclude that it is always preferable to use \sysname for efficient multitask inference, regardless of the amount of shareable neurons, of the probability of executing each task combination, of the network architecture, or of the pruning method used after merging.

\subsection{Ablation Study}
\label{subsec:additional}
This subsection presents experiments to further understand the effectiveness of \sysname. 

\begin{figure}[t]
\centering
     \subfloat[]{
         \label{fig:shared_neurons}
     	\includegraphics[width=0.47\linewidth]{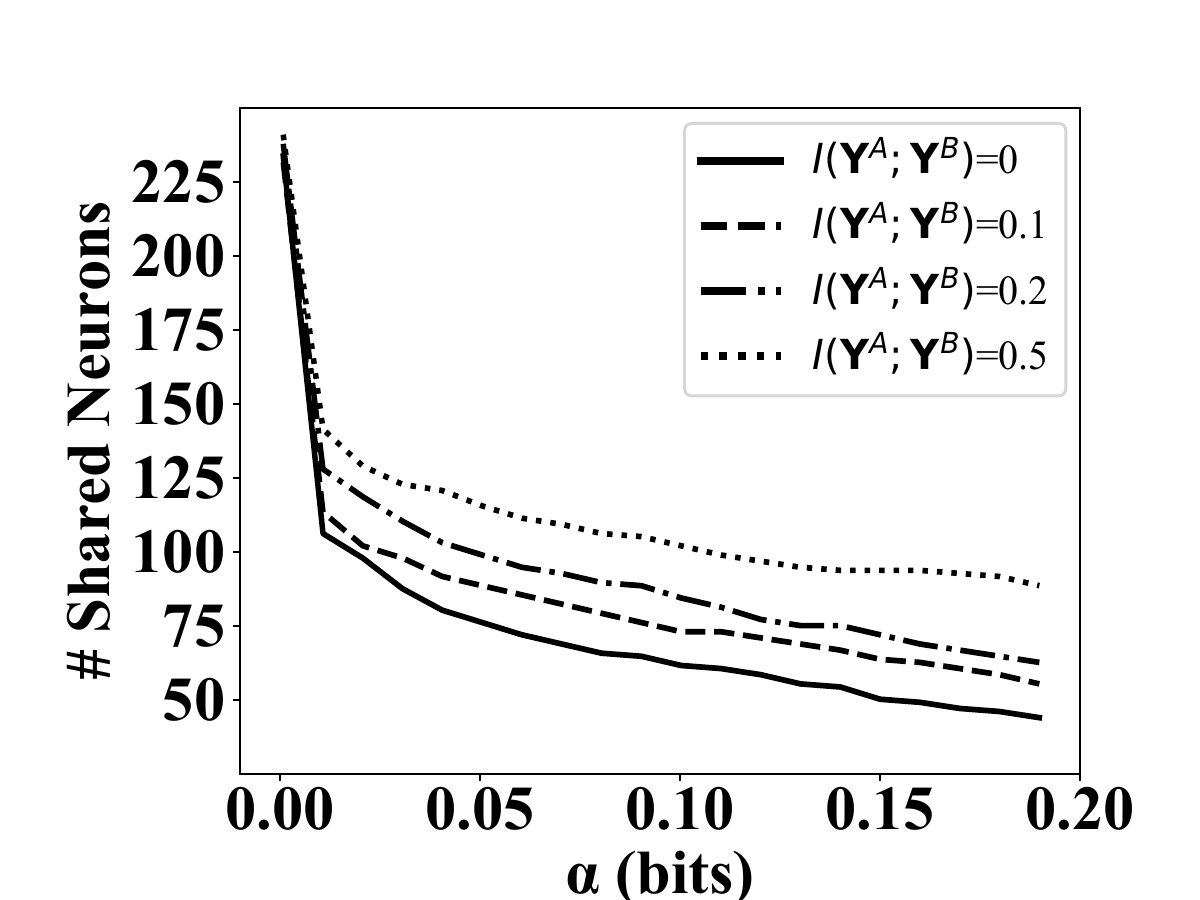}}
     \subfloat[]{
         \label{fig:subtask}
     	\includegraphics[width=0.47\linewidth]{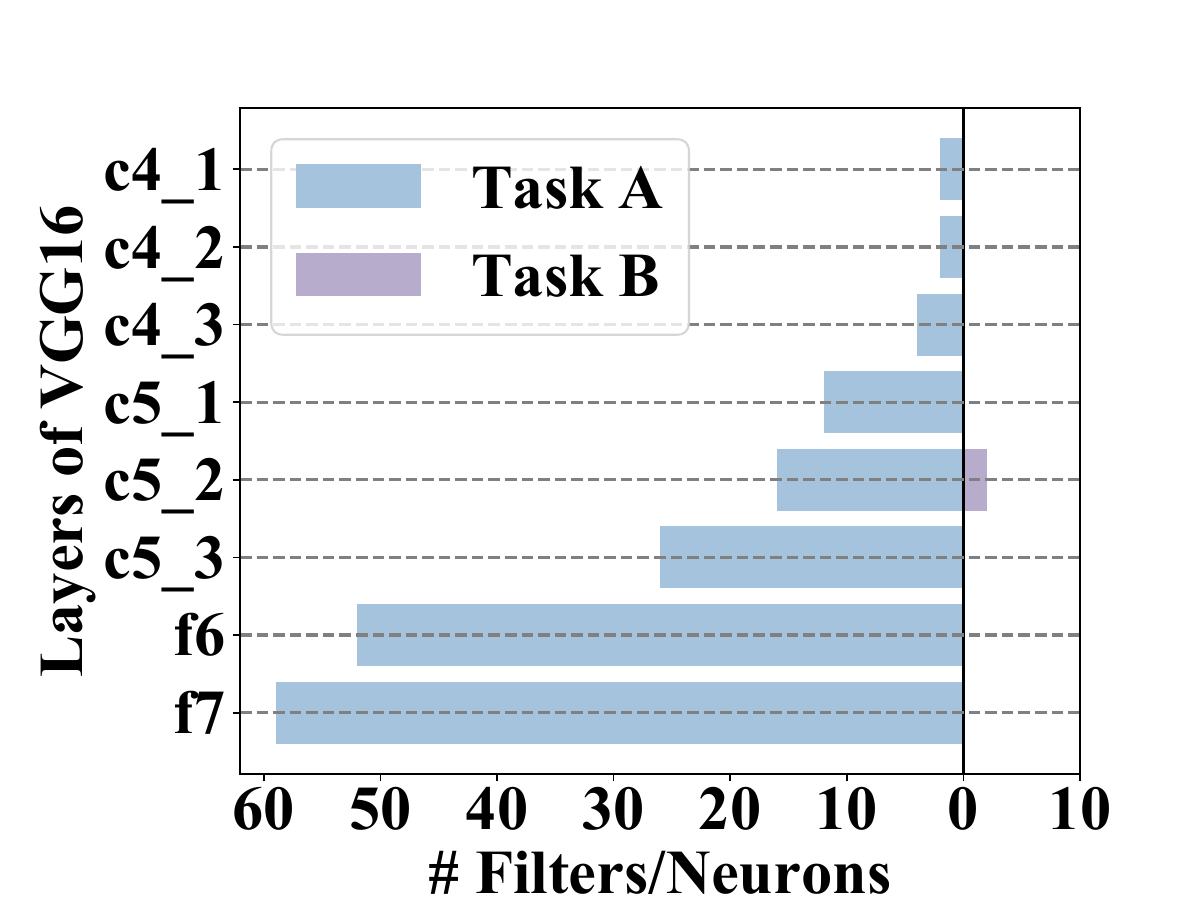}}
     \caption{Ablation studies: (a) Number of shared neurons in layer \texttt{f7} of the four multitask networks constructed with \sysname for different task pairs on LFW dataset, with different tuning parameter $\alpha$. (b) The number of non-shared neurons in $\mathbf{L}'^A_i$ and $\mathbf{L}'^B_i$ in the last eight layers when task $B$ is a sub-task of task $A$. The networks are trained and merged on LFW.} 
     \label{fig:ablation}
 \end{figure}
 
\subsubsection{Impact of Task Relatedness}
\label{appendix:task_relatedness}
This study aims to show the impact of task relatedness on the performance gain \sysname can achieve.
The number of neurons that can be shared among pre-trained networks is related to the relatedness among tasks.
An effective network merging scheme should enforce increasing numbers of shared neurons between tasks with the increase of task relatedness.

\fakeparagraph{Settings}
We consider the 73 labels in LFW as 73 binary classification tasks, and measure the relatedness between each task pair by $I(\mathbf{Y}^A;\mathbf{Y}^B)$.
We then pick four pairs of tasks with $I(\mathbf{Y}^A;\mathbf{Y}^B)\approx 0$, $0.1$, $0.2$ and $0.5$ bits, train four pairs of single-task VGG-16's on them, and construct four multitask networks using \sysname.

\fakeparagraph{Results}
\figref{fig:shared_neurons} plots the number of shared neurons in layer \texttt{f7} of these four multitask networks with different tuning threshold $\alpha$.
The multitask networks for tasks pairs with higher correlation always share neurons.
Hence, \sysname can share an increasing number of neurons between tasks with the increase of task relatedness.

\subsubsection{Case Study: Task Inclusion}
\label{appendix:task_inclusion}
This study aims to validate the effectiveness of \sysname in an extreme yet common case of task relatedness where task $B$ is a sub-task of task $A$.
Ideally, when the mutual information is precisely estimated and true largest sets of task-exclusive neurons are selected, \sysname should effectively pick out only task-$A$-exclusive neurons.

\fakeparagraph{Settings}
We pick 30 labels in LFW as task $A$ and 15 of them as task $B$.
Hence task $A$ includes task $B$.
We train two single-task VGG-16's on these two tasks separately and then merge them by \sysname.

\fakeparagraph{Results}
\figref{fig:subtask} shows the number of non-shared neurons in $\mathbf{L}'^A_i$ and $\mathbf{L}'^B_i$ in the last eight layers of the merged network (the previous layers have exclusively shared neurons).
Almost no neurons are selected for $\mathbf{L}'^B_i$ by Algorithm \ref{Algo}, validating its effectiveness.

\section{Conclusion}
\label{sec:conclusion}

In this paper, we investigate network merging schemes for efficient multitask inference.
Given a set of single-task networks pre-trained for individual tasks, we aim to construct a multitask network such that applying existing network pruning methods on it can minimise the computation cost when performing any subset of tasks.
We theoretically identify the conditions on the multitask network, and design Pruning-Aware Merging (\sysname), a heuristic network merging scheme to construct such a multitask network.
The merged multitask network can then be effectively pruned by existing network pruning methods.
Extensive evaluations show that pruning a multitask network constructed by \sysname achieves low computation cost when performing any subset of tasks in the network.

\bibliographystyle{ACM-Reference-Format}
\bibliography{cites_short}

\appendix

\section*{Appendix}
\label{sec:appendix}

\section{Proofs}
\label{appendix:proofs}

\subsection{Proof of Problem 1 in \secref{subsec:redundancy_multi}}
\label{appendix:problem_1}
Problem 1 occurs because of the lemma below.
\begin{lemma}
Reducing $\mathcal{R}_B(\widetilde{\mathbf{L}}_i^B)$ may decrease $I(\widetilde{\mathbf{L}}_i^A;\mathbf{Y}^A)$.
\end{lemma}
\begin{proof}
We decompose $I(\widetilde{\mathbf{L}}_i^A;\mathbf{Y}^A)$:
\begin{equation}\label{eqt:eq_1}
\begin{aligned}
& I(\widetilde{\mathbf{L}}_i^A;\mathbf{Y}^A)  =  I(\mathbf{L}_i'^A;\mathbf{Y}^A) + \\
& I(\mathbf{L}_i'^{A,B};\mathbf{Y}^A|\mathbf{L}'^A_i, \mathbf{Y}^B) + I(\mathbf{L}_i'^{A,B};\mathbf{Y}^A; \mathbf{Y}^B|\mathbf{L}'^A_i)
\end{aligned}
\end{equation}
where $I(A;B;C)=I(A;B)-I(A;B|C)$ is the \textit{co-information}~\cite{bib:ICA03:Bell}.
From Definition \ref{def:single}, we have:
\begin{align}
\mathcal{R}_B(\widetilde{\mathbf{L}}^B_i) = \medop{\sum_{\widetilde{L}^B_{i,j}\in \widetilde{\mathbf{L}}^B_i}}H(\widetilde{L}^B_{i,j})-I(\widetilde{\mathbf{L}}^B_i;\mathbf{Y}^B) \nonumber\\
=  \medop{\sum_{\widetilde{L}^B_{i,j}\in \widetilde{\mathbf{L}}^B_i}}H(\widetilde{L}^B_{i,j}) - H(\widetilde{\mathbf{L}}^B_i) + H(\widetilde{\mathbf{L}}^B_i|\mathbf{Y}^B)
\end{align}

For the last term, we have: 
\begin{align}
& H(\widetilde{\mathbf{L}}^B_i|\mathbf{Y}^B) \nonumber \\
= & H(\mathbf{L}'^B_i, \mathbf{L}_i'^{A,B}|\mathbf{Y}^B) \\
= & H(\mathbf{L}_i'^{A,B}|\mathbf{Y}^B) + H(\mathbf{L}'^B_i|\mathbf{L}_i'^{A,B}, \mathbf{Y}^B) \\
= & I(\mathbf{L}_i'^{A,B}; \mathbf{Y}^A|\mathbf{Y}^B) + H(\mathbf{L}_i'^{A,B}|\mathbf{Y}^A, \mathbf{Y}^B) + H(\mathbf{L}'^B_i|\mathbf{L}_i'^{A,B}, \mathbf{Y}^B) \\
= & I(\mathbf{L}_i'^{A,B};\mathbf{Y}^A|\mathbf{L}'^A_i, \mathbf{Y}^B)+ 
 I(\mathbf{L}_i'^{A,B};\mathbf{Y}^A; \mathbf{L}'^A_i| \mathbf{Y}^B) + \nonumber \\
& H(\mathbf{L}_i'^{A,B}|\mathbf{Y}^A, \mathbf{Y}^B) + 
H(\mathbf{L}'^B_i|\mathbf{L}_i'^{A,B}, \mathbf{Y}^B)
\end{align}
Hence, $H(\widetilde{\mathbf{L}}^B_i|\mathbf{Y}^B)$ includes $I(\mathbf{L}_i'^{A,B};\mathbf{Y}^A|\mathbf{L}'^A_i, \mathbf{Y}^B)$.
Reducing $\mathcal{R}_B(\widetilde{\mathbf{L}}_i^B)$ may decrease $I(\widetilde{\mathbf{L}}_i^A;\mathbf{Y}^A)$.
\end{proof}

\subsection{Proof of Theorem~\ref{theorem:PAM}}
\label{appendix:proof_1}

\begin{proof}
The proof shows the conditions in Theorem~\ref{theorem:PAM} solve \textit{(i)} Problem 1 in \secref{subsec:redundancy_multi} and \textit{(ii)} Problem 2 in \secref{subsec:redundancy_multi}.

\fakeparagraph{Solving Problem 1 in \secref{subsec:redundancy_multi}}
From \eqref{eqt:eq_1} we have the following if $I(\mathbf{L}_i'^{A,B};\mathbf{Y}^A|\mathbf{L}_i'^A,\mathbf{Y}^B) =0$:
\begin{equation} 
I(\widetilde{\mathbf{L}}_i^A;\mathbf{Y}^A)  = I(\mathbf{L}_i'^A;\mathbf{Y}^A) + I(\mathbf{L}_i'^{A,B};\mathbf{Y}^A; \mathbf{Y}^B|\mathbf{L}'^A_i)
\end{equation}

$\mathbf{L}_i'^A$ is not in $\widetilde{\mathbf{L}}^B_i$.
Hence $I(\mathbf{L}_i'^A;\mathbf{Y}^A)$ is unaffected when $\mathcal{R}_B(\widetilde{\mathbf{L}}^B_i)$ is reduced. 
$I(\mathbf{L}_i'^{A,B};\mathbf{Y}^A; \mathbf{Y}^B|\mathbf{L}'^A_i)$ is included in $I(\widetilde{\mathbf{L}}^B_i;\mathbf{Y}^B)$. 
Thus minimising $\mathcal{R}_B(\widetilde{\mathbf{L}}_i^B) - \tilde{\xi}^B_i \cdot I(\widetilde{\mathbf{L}}_i^B;\mathbf{Y}^B)$ will not reduce $I(\mathbf{L}_i'^{A,B};\mathbf{Y}^A; \mathbf{Y}^B|\mathbf{L}'^A_i)$ with a proper $\tilde{\xi}^B_i$. 
All still hold if we swap $A$ and $B$ in the above equations.
Consequently, if $I(\mathbf{L}_i'^{A,B};\mathbf{Y}^A|\mathbf{L}_i'^A,\mathbf{Y}^B) = I(\mathbf{L}_i'^{A,B};\mathbf{Y}^B|\mathbf{L}_i'^B,\mathbf{Y}^A)$ = $0$, the first two objectives in optimisation problem \eqref{eqt:multi_pruning} become non-conflicting.

\fakeparagraph{Solving Problem 2 in \secref{subsec:redundancy_multi}}
We first decompose $\mathcal{R}_{A,B}(\mathbf{L}_i^{A,B})$ as in \tabref{tab:decomposition}.
Then from \eqref{eq:decomposed}, we have
\begin{align}
    &\mathcal{R}_{A,B}(\mathbf{L}^{A,B}) - \big(\mathcal{R}_{A}(\widetilde{\mathbf{L}}_i^A) + \mathcal{R}_{B}(\widetilde{\mathbf{L}}_i^B)\big)\nonumber\\
    \leq& I(\widetilde{\mathbf{L}}_i^A;\widetilde{\mathbf{L}}_i^B;\{\mathbf{Y}^A,\mathbf{Y}^B\}) - \medop{\sum_{L_{i,j}\in \mathbf{L}_i'^{A,B}}} H(L_{i,j})\\
    \leq& I(\widetilde{\mathbf{L}}_i^A;\widetilde{\mathbf{L}}_i^B) - \medop{\sum_{L_{i,j}\in \mathbf{L}_i'^{A,B}}} H(L_{i,j})\\
    =& I(\mathbf{L}_i'^A,\mathbf{L}_i'^{A,B}; \mathbf{L}_i'^B,\mathbf{L}_i'^{A,B}) - \medop{\sum_{L_{i,j}\in \mathbf{L}_i'^{A,B}}} H(L_{i,j}) \label{eqt:synergy_1}\\
    \leq& I(\mathbf{L}_i'^A; \mathbf{L}_i'^B) + H(\mathbf{L}_i'^{A,B})-  \medop{\sum_{L_{i,j}\in \mathbf{L}_i'^{A,B}}} H(L_{i,j}) \label{eqt:synergy_2}\\
    \leq& I(\mathbf{L}_i'^A; \mathbf{L}_i'^B)
\end{align}

Further,
\begin{align}
     & I(\mathbf{L}_i'^A;\mathbf{L}_i'^B) \nonumber \\
     =&I(\mathbf{L}_i'^A;\mathbf{L}_i'^B;\mathbf{Y}^A;\mathbf{Y}^B)+ I(\mathbf{L}_i'^A;\mathbf{L}_i'^B;\mathbf{Y}^A|\mathbf{Y}^B) + I(\mathbf{L}_i'^A;\mathbf{L}_i'^B|\mathbf{Y}^A)\\
    \leq &I(\mathbf{L}_i'^A;\mathbf{L}_i'^B;\mathbf{Y}^A;\mathbf{Y}^B)
    +H(\mathbf{L}_i'^A|\mathbf{Y}^A)+H(\mathbf{L}_i'^B|\mathbf{Y}^B)\\
    \leq &I(\mathbf{L}_i'^A;\mathbf{L}_i'^B;\mathbf{Y}^A;\mathbf{Y}^B)+\mathcal{R}_{A}(\widetilde{\mathbf{L}}_i^A) + \mathcal{R}_{B}(\widetilde{\mathbf{L}}_i^B)
\end{align}
This is a loose upper bound. 
However, since $\mathcal{R}_{A,B}(\mathbf{L}^{A,B})$, $\mathcal{R}_{A}(\widetilde{\mathbf{L}}_i^A)$ and $\mathcal{R}_{B}(\widetilde{\mathbf{L}}_i^B)$ are lower bounded by $0$, it suffices to show that when $I(\mathbf{L}_i'^A;\mathbf{L}_i'^B;\mathbf{Y}^A;\mathbf{Y}^B) = 0$, minimising $\mathcal{R}_{A}(\widetilde{\mathbf{L}}_i^A)$ and $\mathcal{R}_{B}(\widetilde{\mathbf{L}}_i^B)$ will minimise $\mathcal{R}_{A,B}(\mathbf{L}^{A,B})$.

In summary, when
\begin{equation}
    \begin{aligned}
     I(\mathbf{L}_i'^A;\mathbf{L}_i'^B;\mathbf{Y}^A;\mathbf{Y}^B) = 0 \\ I(\mathbf{L}_i'^{A,B};\mathbf{Y}^A|\mathbf{L}_i'^A,\mathbf{Y}^B) = 0 \\ I(\mathbf{L}_i'^{A,B};\mathbf{Y}^B|\mathbf{L}_i'^B,\mathbf{Y}^A) = 0
    \end{aligned}
 \end{equation}
 the optimisation problem \eqref{eqt:multi_pruning} is reduced to two non-conflicting optimisation problems \eqref{eqt:2opt}.
\end{proof}

\begin{table}[t]
\caption{Decomposition of $\mathcal{R}_{A,B}(\mathbf{L}_i^{A,B})$.}
\label{tab:decomposition}
\hrule
\begin{align}
    &\medmath{\mathcal{R}_{A,B}(\mathbf{L}_i^{A,B})} \nonumber \\
    =&\medmath{\medop{\sum_{L^{A,B}_{i,j}\in \mathbf{L}^{A,B}_i}} H(L^{A,B}_{i,j}) 
    - H(\widetilde{\mathbf{L}}_i^A,\widetilde{\mathbf{L}}_i^B)
    +H(\widetilde{\mathbf{L}}_i^A,\widetilde{\mathbf{L}}_i^B|\mathbf{Y}^A, \mathbf{Y}^B)} \\
    =&\medmath{\medop{\sum_{L^{A}_{i,j}\in \widetilde{\mathbf{L}}_i^A}} H(L^{A}_{i,j}) -
    I(\widetilde{\mathbf{L}}_i^A;\mathbf{Y}^A, \mathbf{Y}^B) 
    + \medop{\sum_{L^{B}_{i,j}\in \widetilde{\mathbf{L}}_i^B}} H(L^{B}_{i,j}) -
    I(\widetilde{\mathbf{L}}_i^B;\mathbf{Y}^A, \mathbf{Y}^B)} \nonumber\\
    & \medmath{+I(\widetilde{\mathbf{L}}_i^A;\widetilde{\mathbf{L}}_i^B; \mathbf{Y}^A, \mathbf{Y}^B) - 
    \medop{\sum_{L_{i,j}\in \mathbf{L}_i'{A,B}}} H(L_{i,j})}\\
    =&\medmath{\medop{\sum_{L^{A}_{i,j}\in \widetilde{\mathbf{L}}_i^A}}H(L^{A}_{i,j}) -
    I(\widetilde{\mathbf{L}}_i^A;\mathbf{Y}^A) - 
    I(\widetilde{\mathbf{L}}_i^A;\mathbf{Y}^B | \mathbf{Y}^A) 
    + \medop{\sum_{L^{B}_{i,j}\in \widetilde{\mathbf{L}}_i^B}}H(L^{B}_{i,j}) -
    I(\widetilde{\mathbf{L}}_i^B;\mathbf{Y}^B)} \nonumber \\
    & \medmath{-I(\widetilde{\mathbf{L}}_i^B;\mathbf{Y}^A | \mathbf{Y}^B) +I(\widetilde{\mathbf{L}}_i^A;\widetilde{\mathbf{L}}_i^B; \mathbf{Y}^A, \mathbf{Y}^B) - 
    \medop{\sum_{L_{i,j}\in \mathbf{L}_i'{A,B}}}H(L_{i,j})}\\
    =& \medmath{\mathcal{R}_{A}(\widetilde{\mathbf{L}}_i^A) + \mathcal{R}_{B}(\widetilde{\mathbf{L}}_i^B) -
    I(\widetilde{\mathbf{L}}_i^A;\mathbf{Y}^B | \mathbf{Y}^A)} \nonumber \\
    &\medmath{-I(\widetilde{\mathbf{L}}_i^B;\mathbf{Y}^A | \mathbf{Y}^B) 
    + I(\widetilde{\mathbf{L}}_i^A;\widetilde{\mathbf{L}}_i^B; \{ \mathbf{Y}^A, \mathbf{Y}^B\} ) -
    \medop{\sum_{L_{i,j}\in \mathbf{L}_i'{A,B}}}H(L_{i,j})} \label{eq:decomposed}
\end{align}
\hrule
\end{table}

\subsection{Proof of Theorem~\ref{theorem:approx}}
\label{appendix:proof_2}
\begin{proof}
First, for co-information between four random variables, we have from~\cite{bib:ICA03:Bell}:
    \begin{equation}
        \label{eqt:heu_1}
        0\leq I(\mathbf{L}_i'^A;\mathbf{L}_i'^B;\mathbf{Y}^A;\mathbf{Y}^B) 
        \leq  \min \{I(\mathbf{L}_i'^A;\mathbf{Y}^B), I(\mathbf{L}_i'^B;\mathbf{Y}^A)\}
    \end{equation}
Therefore, the first condition in Theorem~\ref{theorem:PAM}, \ie $I(\mathbf{L}_i'^A;\mathbf{L}_i'^B;\mathbf{Y}^A;\mathbf{Y}^B)$ = $0$, is achieved by minimising $I(\mathbf{L}_i'^A;\mathbf{Y}^B)$ and $I(\mathbf{L}_i'^B;\mathbf{Y}^A)$ to $0$.

For the second condition in Theorem~\ref{theorem:PAM}, \ie $I(\mathbf{L}_i'^{A,B};\mathbf{Y}^A|\mathbf{L}_i'^A,\mathbf{Y}^B)=0$, we have:
\begin{align}
& I(\mathbf{L}_i'^{A,B};\mathbf{Y}^A|\mathbf{L}_i'^A,\mathbf{Y}^B) \nonumber \\
\leq &H(\mathbf{Y}^A|\mathbf{L}_i'^A,\mathbf{Y}^B) \\
= &H(\mathbf{Y}^A|\mathbf{Y}^B) - I(\mathbf{Y}^A;\mathbf{L}_i'^A) 
+ I(\mathbf{Y}^A;\mathbf{L}_i'^A;\mathbf{Y}^B)\\
\leq &H(\mathbf{Y}^A|\mathbf{Y}^B) - I(\mathbf{Y}^A;\mathbf{L}_i'^A) + I(\mathbf{L}_i'^A;\mathbf{Y}^B)
\end{align}
Given $A$ and $B$, $H(\mathbf{Y}^A|\mathbf{Y}^B)$ is constant. 
The second condition in Theorem~\ref{theorem:PAM} is achieved by minimising $I(\mathbf{L}_i'^A;\mathbf{Y}^B)$ to $0$ and maximising $I(\mathbf{Y}^A;\mathbf{L}_i'^A)$ to $H(\mathbf{Y}^A|\mathbf{Y}^B)$.

The same holds if we swap $A$ and $B$.
The third condition in Theorem~\ref{theorem:PAM}, \ie $I(\mathbf{L}_i'^{A,B};\mathbf{Y}^B|\mathbf{L}_i'^B,\mathbf{Y}^A) = 0$, is achieved by minimising $I(\mathbf{L}_i'^B;\mathbf{Y}^A)$ and maximising $I(\mathbf{Y}^B;\mathbf{L}_i'^B)$.
\end{proof}

\section{Detailed Dataset Setup}
\label{appendix:dataset}

\fakeparagraph{Fashion-MNIST}
The Fashion-MNIST dataset\footnote{https://github.com/f-rumblefish/Multi-Label-Fashion-MNIST} contains $8000$ training images and $2000$ test images with a resolution of $496 \times 124$.
Each image has four fashion product images randomly selected from Fashion-MNIST~\cite{bib:arXiv17:Xiao}.
The 10 categories of fashion products is considered as 10 binary classification problem, and we divide them into two groups (5/5) to form task $A$ and $B$.
On each task we train a LeNet-5, a commonly used architecture for Fashion-MNIST.

\fakeparagraph{CelebA}
\label{sec:exp_detail_celeba}
The CelebA dataset\footnote{http://mmlab.ie.cuhk.edu.hk/projects/CelebA.html} contains over $200$ thousand celebrity face images labelled with $40$ attributes.
The $40$ attributes is divided into two groups ($20$/$20$) to form task $A$ and $B$.
The dataset is divided into training and test sets containing 80\% and 20\% of the samples. 
The input picture resolution is resized to $72\times72$.
On each task we train slightly modified VGG-16 models, a commonly used single-task network architecture on CelebA.
The width of the fully connected layers in VGG-16 is changed to 512.
The convolutional layers are initialised with weights pre-trained for imdb-wiki~\cite{bib:IJCV18:Rothe}, and use the same pre-processing steps.

\fakeparagraph{LFW}
The Labeled Faces in the Wild (LFW) dataset\footnote{http://vis-www.cs.umass.edu/lfw/} contains over 13,000 face photographs collected from the web.
Each face photo is associated with 73 attributes~\cite{bib:ICCV09:Kumar}.
We randomly split the 73 labels in the LFW dataset into four groups with 15 labels each and one group with 13 labels. 
Each group of labels forms a single task.
The dataset is divided into training and test sets containing 80\% and 20\% of the samples. 
Same as in CelebA, the input picture resolution is resized to $72\times72$.
On each task we train slightly modified VGG-16 models, a commonly used single-task network architecture on LFW.
The width of the fully connected layers in VGG-16 is changed to 128.
The convolutional layers are initialised with weights pre-trained for imdb-wiki~\cite{bib:IJCV18:Rothe}, and use the same pre-processing steps.

\tabref{tab:pretrained} summarises the inference accuracy and FLOPs of the \textit{pre-trained} single-task networks. 

\begin{table}[t]
    \caption{Test accuracy and computation cost of pre-trained single-task networks.}
    \label{tab:pretrained}
    \scriptsize
    \centering
    \begin{tabular}{@{}llll@{}}
    \toprule
    Model/Dataset & Task & Accuracy & FLOPs ($\times10^6$) \\ \midrule
    \multirow{2}{*}{LeNet-5/Fashion-MNIST}  & A & 96.05\% & 106.42   \\ \cmidrule(l){2-4} 
                                            & B & 96.37\% & 106.42   \\ \midrule
    \multirow{2}{*}{VGG-16/CelebA}          & A & 90.28\% & 3112.20   \\ \cmidrule(l){2-4} 
                                            & B & 89.03\% & 3112.20   \\ \midrule 
    \multirow{5}{*}{VGG-16/LFW}             & A & 90.23\% & 3110.12   \\ \cmidrule(l){2-4}
                                            & B & 84.15\% & 3110.12   \\ \cmidrule(l){2-4} 
                                            & C & 85.03\% & 3110.12   \\ \cmidrule(l){2-4} 
                                            & D & 86.62\% & 3110.12   \\ \cmidrule(l){2-4} 
                                            & E & 87.44\% & 3110.12   \\ \midrule
    \multirow{2}{*}{ResNet-18/CelebA}       & A & 90.56\% & 994.00   \\ \cmidrule(l){2-4} 
                                            & B & 88.91\% & 994.00   \\ \midrule 
    \multirow{2}{*}{ResNet-34/CelebA}       & A & 90.42\% & 1115.06   \\ \cmidrule(l){2-4} 
                                            & B & 88.70\% & 1115.06   \\ \bottomrule
    \end{tabular}
\end{table}

\section{Visualisation of Algorithm~\ref{Algo}}
\label{appendix:alpha}

\begin{figure}[t]
\centering
\includegraphics[width=0.24\textwidth]{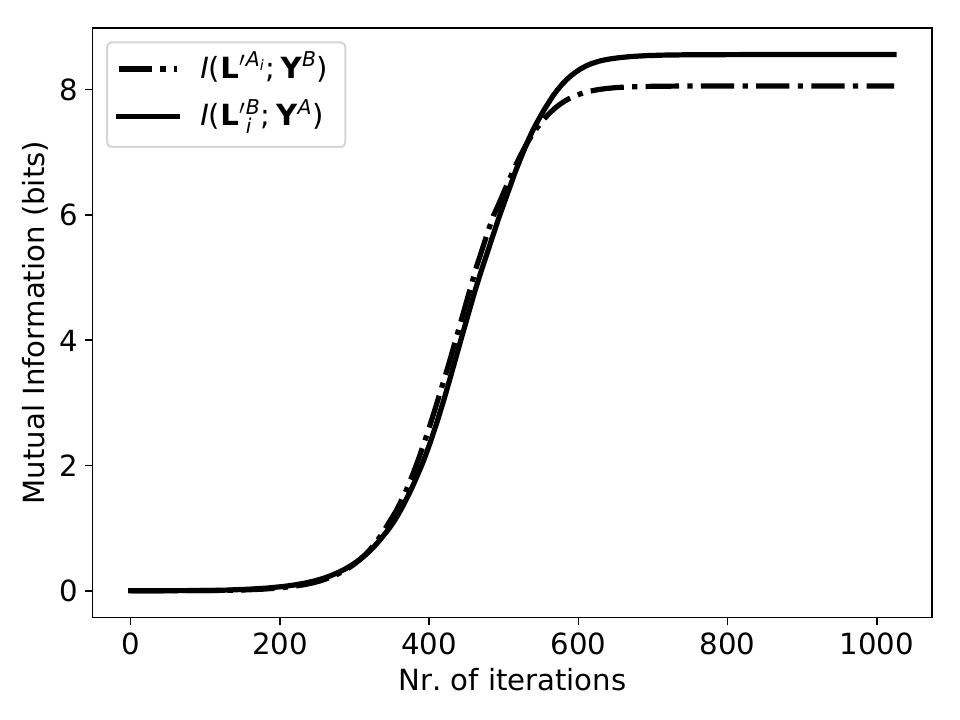}
\caption{Iterations of Line 19-22 and 24-27 in Algorithm~\ref{Algo}. The shown example is on the \texttt{f7} layer of the VGG-16 networks trained and merged on CelebA.}
\label{fig:alpha}
\end{figure}

\figref{fig:alpha} illustrates two iterations of Line 19-22 and 24-27 in Algorithm~\ref{Algo} by showing $I(\mathbf{L}'^A_i; \mathbf{Y}^B)$ and $I(\mathbf{L}'^B_i; \mathbf{Y}^A)$ against the number of iterations.
Here we use the \texttt{f7} layer of VGG-16 trained and merged for CelebA dataset as an example.
The tuning parameter $\alpha$ is set to infinitely large in order to show all the possible cases of the iterations.
From \figref{fig:alpha}, we can observe three phases:
\begin{enumerate}
    \item 
    In the first phase, $I(\mathbf{L}'^A_i; \mathbf{Y}^B)$ and $I(\mathbf{L}'^B_i; \mathbf{Y}^A)$ remains small, indicating that the selected $\mathbf{L}'^A_i$ and $\mathbf{L}'^B_i$ provides little information about the other task.
    \item 
    In the second phase, $I(\mathbf{L}'^A_i; \mathbf{Y}^B)$ and $I(\mathbf{L}'^B_i; \mathbf{Y}^A)$ start to increase as it is impossible to add more neurons to $\mathbf{L}'^A_i$ and $\mathbf{L}'^B_i$ while keeping $I(\mathbf{L}'^A_i; \mathbf{Y}^B)$ and $I(\mathbf{L}'^B_i; \mathbf{Y}^A)$ close to zero.
    \item 
    In the third phase, $I(\mathbf{L}'^A_i; \mathbf{Y}^B)$ and $I(\mathbf{L}'^B_i; \mathbf{Y}^A)$ start to saturate as the newly joined neurons contain mostly information already included in existing $\mathbf{L}'^A_i$ and $\mathbf{L}'^B_i$.
\end{enumerate}

In practice, the parameter $\alpha$ tuned as remains small, and the iterations in Algorithm~\ref{Algo} as well as Algorithm~\ref{Algo_extended} usually stop at the end of the first phase or the beginning of the second phase.

\end{document}